\documentclass[10pt,journal,compsoc]{IEEEtran}
\usepackage[nocompress]{cite}
\usepackage[pdftex]{graphicx}
\usepackage{amsmath,amsfonts,mathtools,amssymb}
\usepackage{algorithm,algorithmic}
\usepackage{array}
\usepackage{eqparbox}
\usepackage[pagebackref,breaklinks,colorlinks,draft=False]{hyperref}
\usepackage[caption=false,font=footnotesize,labelfont=sf,textfont=sf]{subfig}
\usepackage{dblfloatfix}
\usepackage{url}
\usepackage[pdftex]{thumbpdf}
\usepackage{nicefrac}
\usepackage{multirow,booktabs}
\usepackage{xcolor}
\usepackage{bm}
\usepackage{tikz}
\usepackage{lipsum}
\usetikzlibrary{shapes,arrows,fit,positioning}

\DeclareMathOperator{\EX}{\mathbb{E}}

\DeclareMathOperator{\N}{\mathcal{N}}
\DeclareMathOperator{\I}{\mathbf{I}}
\DeclareMathOperator{\x}{\mathbf{x}}
\DeclareMathOperator{\xw}{\Tilde{\mathbf{x}}}
\DeclareMathOperator{\X}{\mathbf{X}}
\DeclareMathOperator{\XW}{\Tilde{\mathbf{X}}}

\begin{document}

\title{Restoring Vision in Adverse Weather Conditions with Patch-Based Denoising Diffusion Models}

\author{Ozan~\"{O}zdenizci~and~Robert~Legenstein%
\IEEEcompsocitemizethanks{\IEEEcompsocthanksitem O.~\"{O}zdenizci and R.~Legenstein are with the Institute of Theoretical Computer Science, Graz University of Technology, Graz, Austria.\protect\\
E-mail: \{ozan.ozdenizci,robert.legenstein\}@igi.tugraz.at%
\IEEEcompsocthanksitem O.~\"{O}zdenizci is also affiliated with TU Graz - SAL Dependable Embedded Systems Lab, Silicon Austria Labs, Graz, Austria.}%
}

\markboth{}%
{\"{O}zdenizci and Legenstein: Restoring Vision in Adverse Weather Conditions with Patch-Based Denoising Diffusion Models}

\IEEEtitleabstractindextext{%
\begin{abstract}
Image restoration under adverse weather conditions has been of significant interest for various computer vision applications.
Recent successful methods rely on the current progress in deep neural network architectural designs (e.g., with vision transformers).
Motivated by the recent progress achieved with state-of-the-art conditional generative models, we present a novel patch-based image restoration algorithm based on denoising diffusion probabilistic models.
Our patch-based diffusion modeling approach enables size-agnostic image restoration by using a guided denoising process with smoothed noise estimates across overlapping patches during inference.
We empirically evaluate our model on benchmark datasets for image desnowing, combined deraining and dehazing, and raindrop removal.
We demonstrate our approach to achieve state-of-the-art performances on both weather-specific and multi-weather image restoration, and experimentally show strong generalization to real-world test images.
\end{abstract}
}

\maketitle

\IEEEpeerreviewmaketitle

\IEEEraisesectionheading{\section{Introduction}\label{sec:introduction}}
\label{sec:intro}

\IEEEPARstart{T}{he} restoration of images under adverse impacts of weather conditions such as heavy rain or snow is of wide interest to computer vision research.
At the extreme, observed images to be restored may contain severe weather related obstructions of the true background (e.g., snow flakes, dense hazing effects), causing a well known ill-posed inverse problem where various solutions can be obtained for the unknown ground truth background.
Deep neural networks (DNNs) are shown to excel at such image restoration tasks compared to traditional approaches~\cite{cai2016dehazenet,Fu:2017DDN,liu2018desnownet}, and this success extends with the current progress in DNN architectural designs, e.g., with vision transformers~\cite{liang2021swinir,Zamir2022Restormer}.
State-of-the-art designs have recently shown its effectiveness in low-level weather restoration problems with transformers~\cite{xiao2022image,Valanarasu:2022CVPR} and multi-layer perceptron based models~\cite{Tu:2022}.
Beyond task-specialized solutions, recent work also proposed to tackle this problem for multiple weather corruptions in unified architectures~\cite{Li:2020CVPR,chen2022learning,li2022all,Valanarasu:2022CVPR}.

Earlier deep learning based solutions to adverse weather restoration have extensively explored task-specific generative modeling methods, mainly with generative adversarial networks (GANs) \cite{qian2018attentive,Zhang:2019IDCGAN,li2019heavy}.
In this setting generative models aim to learn the underlying data distribution for cleared image backgrounds, given weather-degraded examples from a training set.
Due to their stronger expressiveness in that sense, generative approaches further accommodate the potential of better generalization to multi-task vision restoration problems.
Along this line, we introduce a novel solution to this problem by using a state-of-the-art conditional generative modeling approach, with denoising diffusion probabilistic models~\cite{Sohl:2015,Ho:2020}.

Denoising diffusion models have recently demonstrated remarkable success in various generative modeling tasks~\cite{Dhariwal:2021,Rombach:2021,ho2022cascaded,saharia2022photorealistic}.
These architectures were however not yet considered for image restoration under adverse weather conditions, or demonstrated to generalize across multiple image restoration problems.
A major obstacle for their usage in image restoration is their architectural constraint that prohibits size-agnostic image restoration, whereas image restoration benchmarks and real-world problems consist of images with various sizes.

We present a novel perspective to the problem of improving vision in adverse weather conditions using denoising diffusion models.
Particularly for image restoration, we introduce a novel patch-based diffusive restoration approach to enable size-agnostic processing.
Our method uses a guided denoising process for diffusion models by steering the sampling process based on smoothed noise estimates for overlapping patches.
Proposed patch-based image processing scheme further introduces a light-weight diffusion modeling approach, and extends practicality of state-of-the-art diffusion models with extensive computational resource demands.
We experimentally use extreme weather degradation benchmarks on removing snow, combined rain with haze, and removal of raindrops obstructing the camera sensor.
We demonstrate our diffusion modeling perspective to excel at several associated problems. 

Our contributions are summarized as follows:
\begin{itemize}
    \item We present a novel patch-based diffusive image restoration algorithm for arbitrary sized image processing with denoising diffusion models.
    \item We empirically demonstrate our approach to achieve state-of-the-art performance on both weather-specific and multi-weather restoration tasks.
    \item We experimentally present strong generalization from synthetic to real-world multi-weather restoration with our generative modeling perspective.
\end{itemize}

\section{Related Work}
\label{sec:background}

\subsection{Diffusion-based Generative Models}

Diffusion based \cite{Sohl:2015} and score-matching based \cite{Hyvarinen:2005,Vincent:2011} generative models recently regained interest with improvements adopted in \textit{denoising diffusion probabilistic models} \cite{Ho:2020,Nichol:2021} and \textit{noise-conditional score networks} \cite{Song:2019,Song:2020}, reaching exceptional image synthesis capabilities \cite{Dhariwal:2021}.
Both approaches relate to a class of generative models that are based on learning to reverse the process of sequentially corrupting data samples with increasing additive noise, until the perturbed distribution matches a standard normal prior.
This is achieved either by optimizing a time-conditional additive noise estimator~\cite{Ho:2020} or a noise conditional score function (i.e., gradient of log-likelihood)~\cite{Song:2019} parameterized by a DNN.
These models are then used for step-wise denoising of samples from a noise distribution, to obtain samples from the data distribution via Langevin dynamics~\cite{Welling:2011}.
Denoising diffusion models were shown to also implicitly learn these score functions at each noise scale, and both methods were later reframed in a unified continuous-time formulation based on stochastic differential equations~\cite{Song:2021}.

Another resembling perspective links \textit{energy-based models} to this class of generative methods~\cite{DuMordatch:2019,Song:2021HowTo}.
Energy-based models estimate an unnormalized probability density defined via the Boltzmann distribution, by optimizing a DNN that represents the energy function.
At test time one can similarly perform Langevin sampling starting from pure noise towards the learned distribution, however this time using the gradient of the energy function.
Notably, energy-based models differ in its training approach which relies on contrastive divergence methods \cite{hinton2002training,tieleman2009using},
whereas diffusion- and score-based models exploit the sequential forward noising (diffusion) scheme to cover a smoother density across isolated modes of the training data distribution.

Recently, diffusion-based conditional generative models have shown state-of-the-art performance in various tasks such as class-conditional data synthesis with classifier guidance~\cite{Dhariwal:2021}, image super-resolution~\cite{Saharia:2021,ho2022cascaded}, image deblurring~\cite{whang2022deblurring}, text-based image synthesis and editing~\cite{Rombach:2021,saharia2022photorealistic}, and general image-to-image translation tasks (e.g., inpainting, colorization)~\cite{Saharia:2021Palette,Choi:2021,Lugmayr:2022}.
Similar conditional generative modeling applications also exist from a score-based modeling perspective~\cite{Meng:2022,chung2022come}.
Notably, Kawar et al.~\cite{Kawar:2022} recently proposed \textit{denoising diffusion restoration models} for general linear inverse image restoration problems, which exploits pre-trained denoising diffusion models for unsupervised posterior sampling.
In contrast to our model, this approach does not perform conditional generative modeling and does not consider image size agnostic restoration.
More generally, diffusion models were so far not considered for image restoration under adverse weather conditions.

\subsection{Image Restoration in Adverse Weather Conditions}
\label{sec:bg_restoration}

The inverse problem of restoring single images by estimating the background scene under weather related foreground degradations is ill-posed.
In this scenario the observed image only contains a mixture of pixel intensities from the weather distortion (e.g., rain streaks) and the background, which can even be fully occluded.
Traditional model-based restoration methods explored various weather distortion characteristic priors to address this problem~\cite{Yang:2020TPAMI}. 

\textbf{Image Deraining \& Dehazing:} Earliest deep learning era breakthroughs extensively studied the problem of image deraining with convolutional neural networks (CNN), see e.g.~the deep detail network~\cite{Fu:2017DDN,Fu:2017TIP}, and the joint rain detection and removal (JORDER) method~\cite{Yang:2017JORDER}.
Following works explored novel mechanisms such as recurrent context aggregation proposed in RESCAN~\cite{li2018recurrent}, or spatial attention maps in SPANet~\cite{Wang:2019SPANet}.
Concurrently popularized GAN based image-to-image translation models (e.g., pix2pix~\cite{isola2017image}, CycleGAN~\cite{zhu2017unpaired}, perceptual adversarial networks~\cite{Wang:2018PAN}) were found successful in modeling underlying image background structures when simply applied to these problems.
This subsequently led to dedicated generative models tailored for weather restoration tasks, such as image deraining conditional GANs~\cite{Zhang:2019IDCGAN}, or conditional variational image deraining~\cite{Du:2020CVID} based on VAEs.
There has been an independent line of work focusing solely on image dehazing \cite{cai2016dehazenet,liu2019griddehazenet,zhao2021refinednet}, where also similar GAN based generative solutions were adopted~\cite{yang2018towards}.
Recently, more challenging natural extensions to this problem were explored, such as heavy rain removal combined with dehazing tasks in a realistic setting by Li et al.~\cite{li2019heavy} via the heavy rain GAN (HRGAN).
Novel solutions introduced hierarchical multi-scale feature extraction and fusion~\cite{Jiang:2020CVPR}, as well as its extension progressive coupled networks (PCNet)~\cite{Jiang:2021TIP} which were shown to outperform several methods on combined deraining and dehazing tasks.
Most recently Zamir et al.~\cite{zamir2021multi} proposed multi-stage progressive image restoration networks with supervised attention modules (MPRNet), which was shown to excel across several general image restoration tasks.

\textbf{Removing Raindrops:} Beyond removal of rain streaks, another natural extension considers removing raindrops that introduce artifacts on the camera sensor.
Originally Qian et al.~\cite{qian2018attentive} presented a dataset on this phenomena, and proposed an Attentive GAN for raindrop removal.
Concurrently Quan et al.~\cite{quan2019deep} proposed an image-to-image CNN with an attention mechanism (RaindropAttn) for the same problem, and Liu et al.~\cite{liu2019dual} demonstrated the effectiveness of dual residual networks (DuRN), a general purpose image restoration model, on this particular task.
Subsequent work focused on restoring multiple degradation effects such as simultaneous removal of raindrops and rain streaks~\cite{Quan:2021CVPR}.
Most recently Xiao et al. proposed an image deraining transformer (IDT)~\cite{xiao2022image} with state-of-the-art results on generating rain-free images for rain streak removal tasks at various severities, and for raindrop removal.

\textbf{Image Desnowing:} One of the earliest deep learning methods for removing snow artifacts from images was proposed by DesnowNet~\cite{liu2018desnownet} with a CNN-based architecture.
Several existing image deraining solutions were later also shown to perform relatively well on this task (e.g., SPANet~\cite{Wang:2019SPANet}, RESCAN~\cite{li2018recurrent}).
Later Chen et al.~\cite{chen2020jstasr} proposed JSTASR which is specifically designed for size and transparency aware snow removal in a unified framework.
Most recently Zhang et al.~\cite{zhang2021deep} proposed a deep dense multi-scale network (DDMSNet) which exploits simultaneous semantic image segmentation and depth estimation mechanisms to improve image desnowing performance, being one of the most effective solutions presented so far.

\textbf{Multi-Weather Restoration:} There have been recent attempts in unifying multiple restoration tasks within single deep learning frameworks, including generative modeling solutions to restore superimposed noise types~\cite{feng2021deep}, restoring test-time unknown mixtures of noise or weather corruptions~\cite{li2022all}, or specifically adverse multi-weather image degradations~\cite{Li:2020CVPR,chen2022learning,Valanarasu:2022CVPR}.
Seminal work by Li et al.~\cite{Li:2020CVPR} in this context proposed the All-in-One unified weather restoration method which utilizes a multi-encoder and decoder architecture and neural architecture search across task-specific optimized encoders.
Most recently Valanarasu et al.~\cite{Valanarasu:2022CVPR} proposed an alternative state-of-the-art solution to this problem with TransWeather, as an end-to-end vision transformer based multi-weather image restoration model.
Notably, to our interest, these two studies~\cite{Li:2020CVPR,Valanarasu:2022CVPR} use the same combination of weather degradation benchmark datasets~\cite{liu2018desnownet,li2019heavy,qian2018attentive}, hence constructing an accumulated line of comparable progress for this research problem.

\section{Adverse Weather Image Restoration with Patch-Based Denoising Diffusion Models}
\label{sec:methods}

\subsection{Denoising Diffusion Probabilistic Models}

Denoising diffusion models~\cite{Sohl:2015,Ho:2020} are a class of generative models that learn a Markov Chain which gradually converts a Gaussian noise distribution into the data distribution that the model is trained on.
The \textit{diffusion process} (i.e., \textit{forward process}) is a fixed Markov Chain that sequentially corrupts the data $\x_0\sim q(\x_0)$ at $T$ diffusion time steps, by injecting Gaussian noise according to a variance schedule $\beta_1,\ldots,\beta_T$:
\begin{equation}
    q(\x_{t}\vert\x_{t-1}) = \mathcal{N}(\x_{t};\sqrt{1-\beta_t}\x_{t-1},\beta_t \I),
    \label{eq:forward}
\end{equation}
\begin{equation}
    q(\x_{1:T}\vert\x_0) = \prod_{t=1}^T q(\x_{t}\vert\x_{t-1}).
\end{equation}
Diffusion models learn to reverse this predefined forward process in~\eqref{eq:forward} utilizing the same functional form.
The \textit{reverse process} defined by the joint distribution $p_{\theta}(\x_{0:T})$ is a Markov Chain with learned Gaussian denoising transitions starting at a standard normal prior $p(\x_T)=\mathcal{N}(\x_T;\mathbf{0},\I)$:
\begin{equation}
    p_{\theta}(\x_{0:T}) = p(\x_{T}) \prod_{t=1}^T p_{\theta}(\x_{t-1}\vert\x_t),
\end{equation}
\begin{equation}
    p_{\theta}(\x_{t-1}\vert\x_t) = \mathcal{N}(\x_{t-1};\bm{\mu}_{\theta}(\x_t,t),\mathbf{\Sigma}_{\theta}(\x_t,t)).
    \label{eq:reverse}
\end{equation}
Here the reverse process is parameterized by a neural network that estimates $\bm{\mu}_{\theta}(\x_t,t)$ and $\mathbf{\Sigma}_{\theta}(\x_t,t)$.
The \textit{forward process} variance schedule $\beta_t$ can be learned jointly with the model or kept constant~\cite{Ho:2020}, ensuring that $\x_T$ approximately follows a standard normal distribution.

The model is trained by optimizing a variational bound on negative data log likelihood $\mathbb{E}_{q(\x_0)}[-\log p_{\theta}(\x_0)]\leq L_{\theta}$, which can be expanded into~\cite{Ho:2020,Dhariwal:2021}:
\begin{equation}
\begin{split}
L_{\theta} = \EX_{q} \Big[ &  \underbrace{D_{\text{KL}}(q(\x_T|\x_0)\,||\,p(\x_T))}_{L_{T}} \underbrace{-\log p_{\theta}(\x_0|\x_1)}_{L_0} \\ & + \sum_{t>1}\underbrace{D_{\text{KL}}(q(\x_{t-1}|\x_t,\x_0)\,||\,p_{\theta}(\x_{t-1}|\x_t))}_{L_{t-1}} \Big].
\label{eq:obj_expanded}
\end{split}
\end{equation}
This loss was shown to be efficiently optimized via stochastic gradient descent over randomly sampled $L_{t-1}$ terms~\cite{Ho:2020}, taking into consideration that we can marginalize the Gaussian diffusion process to sample intermediate $\x_t$ terms directly from clean data $\x_0$ through:
\begin{equation}
q(\x_t\vert\x_0)=\mathcal{N}(\x_t;\sqrt{\bar{\alpha}_t}\x_0,(1-\bar{\alpha}_t)\I),
\end{equation}
which also can be expressed in closed form:
\begin{equation}
\x_t=\sqrt{\bar{\alpha}_t}\x_0+\sqrt{1-\bar{\alpha}_t}\bm{\epsilon}_t,
\label{eq:sampled_xt}
\end{equation}
where $\alpha_t=1-\beta_t$, $\bar{\alpha}_t=\prod_{i=1}^t\alpha_i$, and $\bm{\epsilon}_t\sim\N(\textbf{0},\I)$ has the same dimensionality as data $\x_0$ and latent variables $\x_t$.

Here the $L_{t-1}$ terms in \eqref{eq:obj_expanded} compare the KL divergence between two Gaussians, $p_{\theta}(\x_{t-1}|\x_t)$ from \eqref{eq:reverse} and $q(\x_{t-1}|\x_t,\x_0)$. The latter is the true unknown generative process posterior conditioned on $\x_0$, denoted by:
\begin{equation}
q(\x_{t-1}|\x_t,\x_0)=\N(\x_{t-1};\bm{\Tilde{\mu}}_t(\x_t,\x_0),\Tilde{\beta}_t\I),
\label{eq:true_cond_on_x0}
\end{equation}
where the distribution parameters can be written as:
\begin{equation}
\bm{\Tilde{\mu}}_t=\frac{1}{\sqrt{\alpha_t}} \left(\x_t - \frac{\beta_t}{\sqrt{1-\bar{\alpha}_t}}\bm{\epsilon}_t\right),\;\;  \Tilde{\beta}_t=\frac{(1-\bar{\alpha}_{t-1})}{(1-\bar{\alpha}_t)}\beta_t,
\end{equation}
by incorporating the property~\eqref{eq:sampled_xt} into $\bm{\Tilde{\mu}}_t(\x_t,\x_0)$~\cite{Ho:2020}.
One can either consider fixed reverse process variances for a simple training objective $\mathbf{\Sigma}_{\theta}(\x_t,t)=\mathbf{\sigma}_t^2\I$ (e.g., $\mathbf{\sigma}_t^2=\Tilde{\beta}_t$)~\cite{Ho:2020}, or optimize $\mathbf{\Sigma}_{\theta}(\x_t,t)$ with a hybrid learning objective~\cite{Nichol:2021}.

The overall training objective for the former, when $p_{\theta}(\x_{t-1}\vert\x_t) = \mathcal{N}(\x_{t-1};\bm{\mu}_{\theta}(\x_t,t),\mathbf{\sigma}_t^2\I)$, corresponds to training a network $\bm{\mu}_{\theta}(\x_t,t)$ that predicts $\bm{\Tilde{\mu}}_t$.
Using an alternative reparameterization of the reverse process by:
\begin{equation}
\bm{\mu}_{\theta}(\x_t,t) = \frac{1}{\sqrt{\alpha_t}} \left(\x_t - \frac{\beta_t}{\sqrt{1-\bar{\alpha}_t}}\bm{\epsilon}_{\theta}(\x_t,t)\right),
\label{eq:reparameterization}
\end{equation}
the model can instead be trained to predict the noise vector $\bm{\epsilon}_{\theta}(\x_t,t)$ by optimizing the re-weighted simplified objective:
\begin{equation}
    \mathbb{E}_{\x_0,t,\bm{\epsilon}_t\sim\N(\mathbf{0},\I)}\Big[\vert\vert\bm{\epsilon}_t -  \bm{\epsilon}_{\theta}(\sqrt{\bar{\alpha}_t}\x_0+\sqrt{1-\bar{\alpha}_t}\bm{\epsilon}_t,t)\vert\vert^2 \Big].
    \label{eq:training_obj}
\end{equation}

In this setting we optimize a network that predicts the noise $\bm{\epsilon}_t$ at time $t$, from $\x_t$. Sampling with the learned parameterized Gaussian transitions $p_{\theta}(\x_{t-1}\vert\x_t)$ can then be performed starting from $\x_T\sim\N(\mathbf{0},\I)$ by:
\begin{equation}
    \x_{t-1}=\frac{1}{\sqrt{\alpha_t}}\left(\x_t - \frac{\beta_t}{\sqrt{1-\bar{\alpha}_t}} \bm{\epsilon}_{\theta}(\x_t,t)\right) + \sigma_t\bm{z},
\end{equation}
where $\bm{z}\sim\mathcal{N}(\mathbf{0},\I)$, which resembles one step of sampling via Langevin dynamics~\cite{Welling:2011}.

A large $T$ and small $\beta_t$ for the forward steps allows the assumption that the reverse process becomes close to a Gaussian, which however leads to costly sampling, e.g., when $T=1000$.
The variance schedule is generally chosen to be $\beta_1<\beta_2<\ldots<\beta_T$, leading to larger updates to be performed for noisier samples.
We focus on using a fixed, linearly increasing variance schedule as originally found sufficient in~\cite{Ho:2020}, whereas learning this schedule based on e.g., signal-to-noise ratio estimates~\cite{Kingma:2021} is also possible.

\subsection{Deterministic Implicit Sampling}

Denoising diffusion implicit models~\cite{song2021ddim} present an accelerated deterministic sampling approach for pre-trained diffusion models, which were shown to yield consistent and better quality image samples.
Implicit sampling exploits a generalized non-Markovian forward process formulation:
\begin{equation}
    q_{\lambda}(\x_{1:T}\vert\x_0) = q_{\lambda}(\x_T\vert\x_0)\prod_{t=2}^T q_{\lambda}(\x_{t-1}\vert\x_t,\x_0),
\end{equation}
where we will rewrite the distribution in \eqref{eq:true_cond_on_x0} in terms of a particular choice of its standard deviation $\lambda_t$ as:
\begin{equation}
q_{\lambda}(\x_{t-1}\vert\x_t,\x_0)=\N(\x_{t-1};\bm{\Tilde{\mu}}_t(\x_t,\x_0),\lambda_t^2\I),
\end{equation}
and the mean denoted in terms of the variance as:
\begin{equation}
\bm{\Tilde{\mu}}_t=\sqrt{\bar{\alpha}_{t-1}}\x_0+\sqrt{1-\bar{\alpha}_{t-1}-\lambda_t^2}\cdot\bm{\epsilon}_t,
\end{equation}
by incorporating the property~\eqref{eq:sampled_xt} into $\bm{\Tilde{\mu}}_t(\x_t,\x_0)$.
Here, by setting $\lambda_t^2=\Tilde{\beta}_t$ the forward process becomes Markov and one recovers the original diffusion model formulation described earlier.
Importantly, the training objective~\eqref{eq:training_obj} remains the same, but only embedded non-Markov forward processes are exploited for inference~\cite{song2021ddim}.

A deterministic implicit sampling behavior sets $\lambda_t^2=0$, hence after generating an initial $\x_T$ from the marginal noise distribution sampling becomes deterministic.
We will similarly use our models by setting $\lambda_t^2=0$.
Implicit sampling using a noise estimator network can then be performed by:
\begin{equation}
\begin{split}
\x_{t-1} = & \sqrt{\bar{\alpha}_{t-1}}\left(\frac{\x_t-\sqrt{1-\bar{\alpha}_t}\cdot\bm{\epsilon}_{\theta}(\x_t,t)}{\sqrt{\bar{\alpha}_t}}\right) \\ & + \sqrt{1-\bar{\alpha}_{t-1}}\cdot\bm{\epsilon}_{\theta}(\x_t,t).
\end{split}
\label{eq:ddim}
\end{equation}

During accelerated sampling one only needs a sub-sequence $\tau_1,\tau_2,\ldots,\tau_S$ of the complete $\{1,\ldots,T\}$ timestep indices.
This helps reducing the number of sampling timesteps up to two orders of magnitude.
We determine this sub-sequence by uniformly interleaving from $\{1,\ldots,T\}$:
\begin{equation}
\tau_i = (i-1)\cdot T / S + 1\,,
\end{equation}
which sets $\tau_1=1$ at the final step of reverse sampling.

\subsection{Conditional Diffusion Models}

\begin{figure}[!t]
\centering
\includegraphics[width=0.48\textwidth]{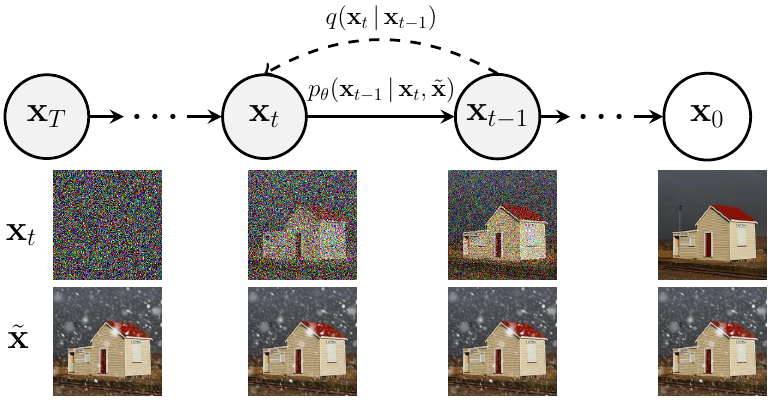}%
\caption{An overview of the forward diffusion (dashed line) and reverse denoising (solid line) processes for a conditional diffusion model.}
\label{fig:diffusion_illustration}
\end{figure}

Conditional diffusion models have shown state-of-the-art image-conditional data synthesis and editing capabilities.
The core idea is to learn a conditional reverse process $p_{\theta}(\x_{0:T}|\xw)$ without modifying the diffusion process $q(\x_{1:T}|\x_0)$ for $\x$, such that the sampled $\x$ has high fidelity to the data distribution conditioned on $\xw$ (see Figure~\ref{fig:diffusion_illustration}).

During training we sample $(\x_0,\xw)\sim q(\x_0,\xw)$ from a paired data distribution (e.g., a clean image $\x_0$ and weather degraded image $\xw$), and learn a conditional diffusion model where we provide $\xw$ as input to the reverse process:
\begin{equation}
p_{\theta}(\x_{0:T}|\xw) = p(\x_{T}) \prod_{t=1}^T p_{\theta}(\x_{t-1}\vert\x_t,\xw).
\end{equation}

Our previous formulation of optimizing a noise estimator network via~\eqref{eq:training_obj} then uses $\bm{\epsilon}_{\theta}(\x_t,\xw,t)$.
For image-based conditioning, inputs $\x$ and $\xw$ are concatenated channel-wise, resulting in six dimensional input image channels.

Note that conditioning the reverse process on $\xw$ maintains its compatibility with implicit sampling.
In this formulation one samples from $\x_{t-1}\sim p_{\theta}(\x_{t-1}\vert\x_t,\xw)$ with:
\begin{equation}
\begin{split}
\x_{t-1} = & \sqrt{\bar{\alpha}_{t-1}}\left(\frac{\x_t-\sqrt{1-\bar{\alpha}_t}\cdot\bm{\epsilon}_{\theta}(\x_t,\xw,t)}{\sqrt{\bar{\alpha}_t}}\right) \\ & + \sqrt{1-\bar{\alpha}_{t-1}}\cdot\bm{\epsilon}_{\theta}(\x_t,\xw,t),
\end{split}
\label{eq:cddim}
\end{equation}
which follows a deterministic reverse path towards $\x_0$ with fidelity to the condition $\xw$, starting from $\x_T\sim\N(\mathbf{0},\I)$.

\begin{figure*}[!t]
\centering
\subfloat[Patch-based diffusive image restoration\label{fig:restoration_illustration}]{\includegraphics[width=0.59\textwidth]{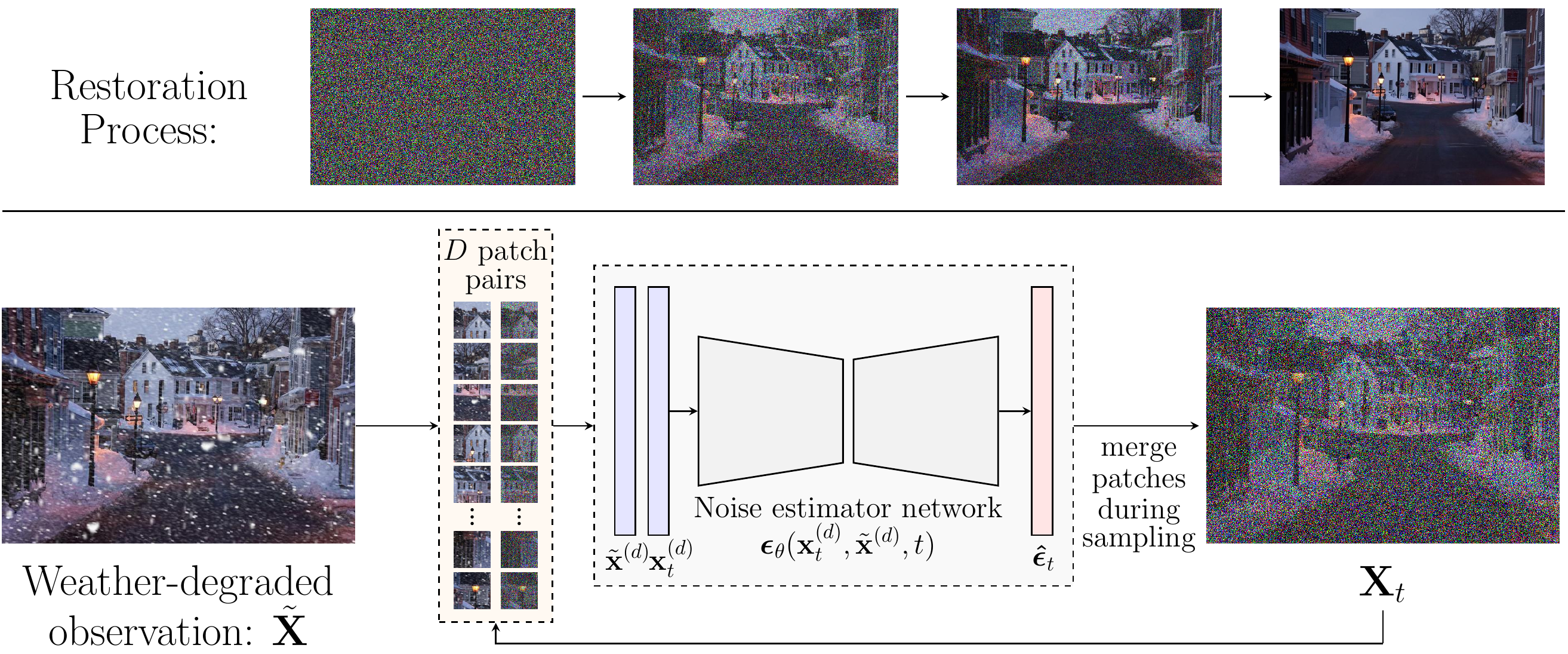}}%
\hspace{0.05cm}
\subfloat[Illustrating sampling for overlapping patches\label{fig:grid_illustration}]{\includegraphics[width=0.4\textwidth]{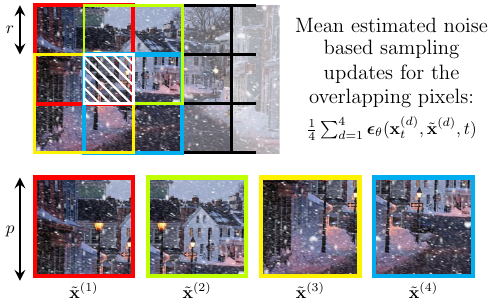}}%
\caption{(a) Illustration of the patch-based diffusive image restoration pipeline detailed in Algorithm~\ref{alg:inference}. (b) Illustrating \textit{mean estimated noise} guided sampling updates for overlapping pixels across patches.
We demonstrate a simplified example where $r=p/2$, and there are only four overlapping patches sharing the grid cell marked with the white border and gratings. In this case, we would perform sampling updates for the pixels in this region based on the mean estimated noise over the four overlapping patches, at each denoising time step $t$.}
\label{fig:patch_based_illustrations}
\end{figure*}

\subsection{Patch-based Diffusive Image Restoration}

Image restoration benchmarks, as well as real world pictures, consist of images with various sizes.
Contrarily, existing generative architectures are mostly tailored for fixed-size image processing.
There has been one recent diffusion modeling work which studied size-agnostic blurred image restoration~\cite{whang2022deblurring}. Their model is optimized using fixed-size patches and then used for deblurring by simply providing arbitrary sized inputs to the model, hence strictly relying on a modified fully-convolutional network architecture. This also leads to high test time computation demands such that the whole image can be processed within memory.
Differently, we decompose images into overlapping fixed-sized patches also at test time and blend them during sampling.

The general idea of patch-based restoration is to operate locally on patches extracted from the image and optimally merge the results.
An important drawback of this approach so far has been that the resulting image can contain merging artifacts from independently restored intermediate results, which was extensively studied in traditional restoration methods~\cite{kervrann2006optimal,zoran2011learning,papyan2015multi}.
We will tackle this problem by guiding the reverse sampling process towards smoothness across neighboring patches, without emerging edge artifacts.

We define the unknown ground truth image of arbitrary size as $\X_0$, the weather-degraded observation as $\XW$, and $\bm{P}_i$ to be a binary mask matrix of same dimensionality as $\X_0$ and $\XW$, indicating the $i$-th $p\times p$ patch location from the image.
Our training approach is outlined in Algorithm~\ref{alg:training}, in which we learn the conditional reverse process:
\begin{equation}
p_{\theta}(\x_{0:T}^{(i)}|\xw^{(i)}) = p(\x_{T}^{(i)}) \prod_{t=1}^T p_{\theta}(\x_{t-1}^{(i)}\vert\x_t^{(i)},\xw^{(i)}),
\end{equation}
with $\x_0^{(i)}=\text{Crop}(\bm{P}_i\circ\X_0)$ and $\xw^{(i)}=\text{Crop}(\bm{P}_i\circ\XW)$ denoting $p\times p$ patches from a training set image pair $(\X_0,\XW)$, where $\text{Crop}(.)$ operation extracts the patch from the location indicated by $\bm{P}_i$.
During training we randomly sample (with uniform probability) the $p\times p$ patch location for $\bm{P}_i$ within the complete range of image dimensions.

Our test time patch-based diffusive image restoration method is illustrated in Figure~\ref{fig:restoration_illustration} and outlined in Algorithm~\ref{alg:inference}.
Firstly, we decompose the image $\XW$ of arbitrary size by extracting all overlapping $p\times p$ patches from a grid-like arranged parsing scheme.
We consider a grid-like arrangement over the complete image where each grid cell contains $r\times r$ pixels ($r<p$), and extract all $p\times p$ patches by moving over this grid with a step size of $r$ in both horizontal and vertical dimensions (see Figure~\ref{fig:grid_illustration} for an illustration).
We define $D$ as the total number of extracted patches, defining a dictionary of overlapping patch locations.

Due to the ill-posed nature of the problem, different restoration estimates for overlapping grid cells will be obtained when performing conditional reverse sampling based on neighboring overlapping patches. 
We alleviate this by performing reverse sampling based on the \textit{mean estimated noise} for each pixel in overlapping patch regions, at any given denoising time step $t$ (see Figure~\ref{fig:grid_illustration}).
Our approach effectively steers the reverse sampling process to ensure higher fidelity across all contributing neighboring patches.
More specifically at each time step $t$ of sampling, (1) we estimate the additive noise for all overlapping patch locations $d\in\{1,\ldots,D\}$ using $\bm{\epsilon}_{\theta}(\x_t^{(d)},\xw^{(d)},t)$, (2) accumulate these overlapping noise estimates at their respective patch locations in a matrix $\bm{\hat{\Omega}}_t$ of same size as the whole image (line 8 in Alg.~\ref{alg:inference}), (3) normalize $\bm{\hat{\Omega}}_t$ based on the number of received estimates for each pixel (line 11 in Alg.~\ref{alg:inference}), (4) perform an implicit sampling update using the smoothed whole-image noise estimate $\bm{\hat{\Omega}}_t$ (line 12 in Alg.~\ref{alg:inference}).

Our method is different from a naive baseline of averaging overlapping final reconstructions after sampling.
Such an approach destroys the local patch distribution fidelity to the learned posterior if applied post-sampling.
(see Section~1.3 of Supplementary Materials for both quantitative evaluations and visual comparisons on this).
Differently from our overlapping patch based guided sampling principle, however in a similar spirit, there are also recently successful image editing methods based on steering the reverse process in the latent space to achieve sampling from a condensed subspace of the learned density~\cite{Choi:2021,Kawar:2022}.

\begin{algorithm}[t]
\caption{Diffusive weather restoration model training}
\label{alg:training}
\begin{algorithmic}[1]
\renewcommand{\algorithmicrequire}{\textbf{Input:}}
\renewcommand{\algorithmicensure}{\textbf{Output:}}
\REQUIRE Clean and weather-degraded image pairs $(\X_0,\XW)$
\REPEAT
\STATE Randomly sample a binary patch mask $\mathbf{P}_i$
\STATE $\x_0^{(i)}=\text{Crop}(\bm{P}_i\circ\X_0)$ and $\xw^{(i)}=\text{Crop}(\mathbf{P}_i\circ\XW)$
\STATE $t\sim \text{Uniform}\{1,\ldots,T\}$
\STATE $\bm{\epsilon}_t\sim\mathcal{N}(\mathbf{0},\I)$
\STATE Perform a single gradient descent step for \\ \qquad $\nabla_{\theta}\vert\vert\bm{\epsilon}_t -  \bm{\epsilon}_{\theta}(\sqrt{\bar{\alpha}_t}\x_0^{(i)}+\sqrt{1-\bar{\alpha}_t}\bm{\epsilon}_t\,,\xw^{(i)},t)\vert\vert^2$
\UNTIL converged
\RETURN $\theta$
\end{algorithmic} 
\end{algorithm}

\begin{algorithm}[t]
\caption{Patch-based diffusive image restoration}
\label{alg:inference}
\begin{algorithmic}[1]
\renewcommand{\algorithmicrequire}{\textbf{Input:}}
\renewcommand{\algorithmicensure}{\textbf{Output:}}
\REQUIRE Weather-degraded image $\XW$, conditional diffusion model $\bm{\epsilon}_{\theta}(\x_t,\xw,t)$, number of implicit sampling steps $S$, dictionary of $D$ overlapping patch locations.
\STATE $\X_{t}\sim\N(\mathbf{0},\mathbf{I})$
\FOR {$i = S,\ldots,1$}
\STATE $t = (i-1)\cdot T / S + 1$
\STATE $t_{\text{next}} = (i-2)\cdot T / S + 1\;$ \textbf{if} $\,i>1$ \textbf{else} $\,0$
\STATE $\bm{\hat{\Omega}}_t=\mathbf{0}$ and $\mathbf{M}=\mathbf{0}$
\FOR {$d = 1,\ldots,D$}
\STATE $\x_t^{(d)}=\text{Crop}(\mathbf{P}_d\circ\X_t)$ and $\xw^{(d)}=\text{Crop}(\mathbf{P}_d\circ\XW)$
\STATE $\bm{\hat{\Omega}}_t = \bm{\hat{\Omega}}_t + \mathbf{P}_d\cdot\bm{\epsilon}_{\theta}(\x_t^{(d)},\xw^{(d)},t)$
\STATE $\mathbf{M} = \mathbf{M} + \mathbf{P}_d$
\ENDFOR
\STATE $\bm{\hat{\Omega}}_t = \bm{\hat{\Omega}}_t\oslash\mathbf{M}$\qquad\quad$\mathbin{/\mkern-4mu/}$\;\,$\oslash$: element-wise division
\STATE $\X_{t}\leftarrow\sqrt{\bar{\alpha}_{t_{\text{next}}}}\left(\frac{\X_t-\sqrt{1-\bar{\alpha}_t}\,\cdot\,\bm{\hat{\Omega}}_t}{\sqrt{\bar{\alpha}_t}}\right) + \sqrt{1-\bar{\alpha}_{t_{\text{next}}}}\cdot\bm{\hat{\Omega}}_t$
\ENDFOR
\RETURN $\X_t$
\end{algorithmic} 
\end{algorithm}

Note that a smaller $r$ increases overlap between patches and hence smoothness, however also the computational burden.
We used $p=64$ or $128$ pixels for $\mathbf{P}_i$, and $r=16$ pixels.
Before processing, we resized whole image dimensions to be multiples of 16 as also conventionally done with vision transformers~\cite{Valanarasu:2022CVPR}.
Here, choosing $r=p$ would construct a set of non-overlapping patches for processing, hence would assume independency across patches during restoration.
However such neighboring patches are clearly not independent in images, and this would lead to a suboptimal approximation with edge artifacts in restored images (see Section~1.3 of Supplementary Materials for ablations).

Our proposed patch-based conditional diffusion modeling approach is task-agnostic, and further extends to simultaneously handling multiple weather corruptions when example image pairs from a mixture of weather degradations are observed at training time, which we will experimentally demonstrate in Section~\ref{sec:results}.
The model is then effectively optimized to estimate the background while restoring images (e.g., approximating the background behind the occlusions from large snowflakes or raindrops) based on a learned mixture of conditional distributions.
Note that while doing so, our model does not require any additional input regarding which task (weather) to consider at training or test time.

\section{Experimental Results}
\label{sec:results}

\subsection{Datasets}
\label{sec:datasets}

We used three standard benchmark image restoration datasets considering adverse weather conditions of snow, heavy rain with haze, and raindrops on the camera sensor.

\textbf{Snow100K \cite{liu2018desnownet}} is a dataset for evaluation of image desnowing models.
It consists of 50,000 training and 50,000 test images split into approximately equal sizes of three Snow100K-S/M/L sub-test sets (16,611/16,588/16,801), indicating the synthetic snow strength imposed via snowflake sizes (light/mid/heavy).
It also contains additional 1,329 real snowy images (Snow100K-Real) to evaluate real-world generalization of models trained with synthetic data.

\textbf{Outdoor-Rain \cite{li2019heavy}} is a dataset of simultaneous rain and fog which exploits a physics-based generative model to simulate not only dense synthetic rain streaks, but also incorporating more realistic scene views, constructing an inverse problem of simultaneous image deraining and dehazing. The Outdoor-Rain training set consists of 9,000 images, and the test set we used, denoted in~\cite{li2019heavy} as Test1, is of size 750 for quantitative evaluations.

\textbf{RainDrop \cite{qian2018attentive}} is a dataset of images with raindrops introducing artifacts on the camera sensor and obstructing the view. It consists of 861 training images with synthetic raindrops, and a test set of 58 images dedicated for quantitative evaluations, denoted in~\cite{qian2018attentive} as RainDrop-A.

\begin{figure*}%
\centering
\subfloat[Image Desnowing \label{tab:snow100k}]{
\scalebox{0.79}{
\begin{tabular}{l c c c c}
\toprule
& \multicolumn{2}{c}{Snow100K-S~\cite{liu2018desnownet}} & \multicolumn{2}{c}{Snow100K-L~\cite{liu2018desnownet}} \\
\cmidrule(l{.5em}r{.5em}){2-3}\cmidrule(l{.5em}r{.5em}){4-5}
& PSNR $\uparrow$ & SSIM $\uparrow$ & PSNR $\uparrow$ & SSIM $\uparrow$ \\
\midrule
SPANet~\cite{Wang:2019SPANet} & 29.92 & 0.8260 & 23.70 & 0.7930 \\
JSTASR~\cite{chen2020jstasr} & 31.40 & 0.9012 & 25.32 & 0.8076 \\
RESCAN~\cite{li2018recurrent} & 31.51 & 0.9032 & 26.08 & 0.8108 \\
DesnowNet~\cite{liu2018desnownet} & 32.33 & 0.9500 & 27.17 & 0.8983 \\
DDMSNet~\cite{zhang2021deep} & 34.34 & 0.9445 & 28.85 & 0.8772 \\
\midrule
\textbf{SnowDiff$_{64}$} & \textbf{36.59} & \textbf{0.9626} & \textbf{30.43} & \textbf{0.9145} \\ 
\textbf{SnowDiff$_{128}$} & \underline{36.09} & \underline{0.9545} & \underline{30.28} & \underline{0.9000} \\
\midrule
\midrule
All-in-One \cite{Li:2020CVPR} & - & - & 28.33 & 0.8820 \\
TransWeather \cite{Valanarasu:2022CVPR} & 32.51 & 0.9341 & 29.31 & 0.8879 \\
\midrule
\textbf{WeatherDiff$_{64}$} & \textbf{35.83} & \textbf{0.9566} &
\textbf{30.09} & \textbf{0.9041} \\
\textbf{WeatherDiff$_{128}$} & \underline{35.02} & \underline{0.9516} & \underline{29.58} & \underline{0.8941} \\
\bottomrule
\end{tabular}}}%
\quad
\subfloat[Image Deraining \& Dehazing\label{tab:outdoorrain}]{
\scalebox{0.79}{
\begin{tabular}{l c c c c}
\toprule
& \multicolumn{2}{c}{Outdoor-Rain~\cite{li2019heavy}}\\
\cmidrule(l{.5em}r{.5em}){2-3}
& PSNR $\uparrow$ & SSIM $\uparrow$ \\
\midrule
CycleGAN~\cite{zhu2017unpaired} & 17.62 & 0.6560 \\
pix2pix~\cite{isola2017image} & 19.09 & 0.7100 \\
HRGAN~\cite{li2019heavy} & 21.56 & 0.8550 \\
PCNet~\cite{Jiang:2021TIP} & 26.19 & 0.9015 \\
MPRNet~\cite{zamir2021multi} & \underline{28.03} & \underline{0.9192} \\
\midrule
\textbf{RainHazeDiff$_{64}$} & \textbf{28.38} & \textbf{0.9320} \\ 
\textbf{RainHazeDiff$_{128}$} & 26.84 & 0.9152 \\
\midrule
\midrule
All-in-One \cite{Li:2020CVPR} & 24.71 & 0.8980 \\
TransWeather \cite{Valanarasu:2022CVPR} & 28.83 & 0.9000 \\
\midrule
\textbf{WeatherDiff$_{64}$} & \underline{29.64} & \textbf{0.9312} \\
\textbf{WeatherDiff$_{128}$} & \textbf{29.72} & \underline{0.9216} \\
\bottomrule
\end{tabular}}}%
\quad
\subfloat[Removing Raindrops\label{tab:raindrop}]{
\scalebox{0.79}{
\begin{tabular}{l c c c c}
\toprule
& \multicolumn{2}{c}{RainDrop~\cite{qian2018attentive}}\\
\cmidrule(l{.5em}r{.5em}){2-3}
& PSNR $\uparrow$ & SSIM $\uparrow$ \\
\midrule
pix2pix~\cite{isola2017image} & 28.02 & 0.8547 \\
DuRN~\cite{liu2019dual} & 31.24 & 0.9259 \\
RaindropAttn~\cite{quan2019deep} & 31.44 & 0.9263 \\
AttentiveGAN~\cite{qian2018attentive} & 31.59 & 0.9170 \\
IDT~\cite{xiao2022image} & 31.87 & 0.9313 \\
\midrule
\textbf{RainDropDiff$_{64}$} & \underline{32.29} & \textbf{0.9422} \\ 
\textbf{RainDropDiff$_{128}$} & \textbf{32.43} & \underline{0.9334} \\
\midrule
\midrule
All-in-One \cite{Li:2020CVPR} & \textbf{31.12} & \underline{0.9268} \\
TransWeather \cite{Valanarasu:2022CVPR} & 30.17 & 0.9157 \\
\midrule
\textbf{WeatherDiff$_{64}$} & \underline{30.71} & \textbf{0.9312} \\ 
\textbf{WeatherDiff$_{128}$} & 29.66 & 0.9225 \\
\bottomrule
\end{tabular}}}
\caption{Quantitative comparisons in terms of PSNR and SSIM (higher is better) with state-of-the-art image desnowing and deraining methods. Above half of the tables show comparisons of our weather-specific SnowDiff$_{p}$, RainHazeDiff$_{p}$ and RainDropDiff$_{p}$ models individually evaluated for each task. Bottom half of the tables show evaluations of our unified multi-weather model WeatherDiff$_{p}$ on all three test sets with respect to All-in-One~\cite{Li:2020CVPR} and TransWeather~\cite{Valanarasu:2022CVPR} multi-weather restoration methods.
Best and second best values are indicated with bold text and underlined text respectively.}%
\label{tab:image_restoration}%
\end{figure*}

\subsection{Diffusion Model Implementations} 
\label{sec:impl_diffusion}

We performed experiments both in weather-specific and multi-weather image restoration settings.
We denote our weather-specific restoration models as \textbf{SnowDiff$_p$}, \textbf{RainHazeDiff$_p$} and \textbf{RainDropDiff$_p$}, and our multi-weather restoration model as \textbf{WeatherDiff$_p$}, with the subscripts denoting the input patch size of the model.
We trained both 64x64 and 128x128 patch size versions of all models.

We used the same diffusion process configuration for all trained models.
We grounded our model selection and hyper-parameters via the definitions used in previous seminal work by~\cite{Ho:2020,song2021ddim}.
The network had a U-Net architecture~\cite{ronneberger2015u} based on WideResNet~\cite{Zagoruyko:2016}, which uses group normalization~\cite{wu2018group} and self-attention blocks at 16x16 feature map resolution~\cite{vaswani2017attention,wang2018non}.
We used input time step embedding for $t$ through sinusoidal positional encoding~\cite{vaswani2017attention} and provided these embeddings as input to each residual block, enabling the model to share parameters across time.
For input image conditioning we channel-wise concatenate the patches $\x_t$ and $\xw$, resulting in six dimensional input image channels (i.e., RGB for both images).
We did not perform task-specific parameter tuning or modifications to the neural network architecture.
Further specifications on the model configurations are provided in Table~1 of Supplementary Materials. Our code is available at: \href{https://github.com/IGITUGraz/WeatherDiffusion}{https://github.com/IGITUGraz/WeatherDiffusion}.

\subsection{Training Specifications}
\label{sec:training_specs}

At each training iteration of 64x64 patch diffusion models, we initially sampled 16 images from the training set and randomly cropped 16 patches of size 64x64 from each, resulting in mini-batches of size 256 patches.
For 128x128 patch diffusion models, we randomly cropped 8 patches from each of the 8 sampled training images per iteration, resulting in mini-batches of size 64.
We used all training set images per epoch for weather-specific restoration.
For WeatherDiff$_p$ we used the curated \textit{AllWeather} dataset from \cite{Valanarasu:2022CVPR}, which has 18,069 samples composed of subsets of training images from Snow100K, Outdoor-Rain and RainDrop, in order to create a balanced training set across three weather conditions with a similar approach to \cite{Li:2020CVPR}.
Our multi-weather models are effectively conditioned to generate the most likely background for any of the three conditions, as we use a mixture of weather degradations in training batches.

We trained all models for 2,000,000 iterations, except for WeatherDiff$_{128}$ which was trained for 2,500,000 iterations due to complexity of this task (see Section~1.1 of Supplementary Materials for an empirical analysis).
We used an Adam optimizer with a fixed learning rate of $0.00002$ without weight decay.
An exponential moving average with a weight of 0.999 was applied during parameter updates, as it was shown to facilitate more stable learning~\cite{Song:2020,Nichol:2021}.

\begin{figure*}[!ht]
\subfloat[Input\label{fig-snow:input}]{%
\begin{tabular}{c}
\includegraphics[width=0.192\textwidth]{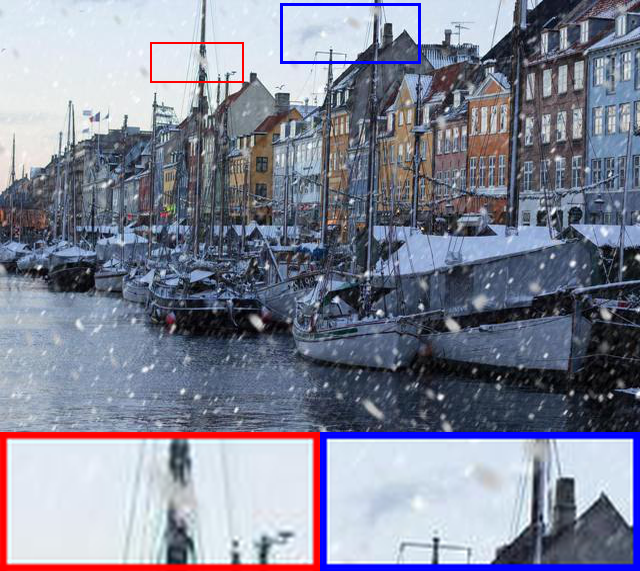} \\
\includegraphics[width=0.192\textwidth]{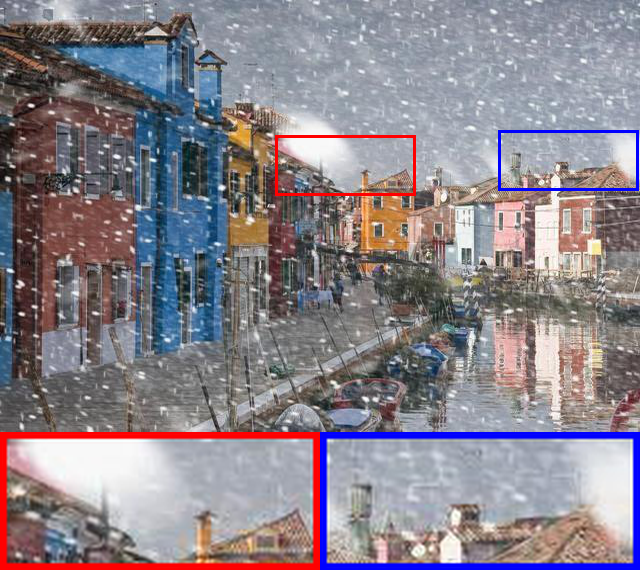}
\end{tabular}%
}\hspace{-.45cm}
\subfloat[DesnowNet~\cite{liu2018desnownet}\label{fig-snow:desnownet}]{%
\begin{tabular}{c}
\includegraphics[width=0.192\textwidth]{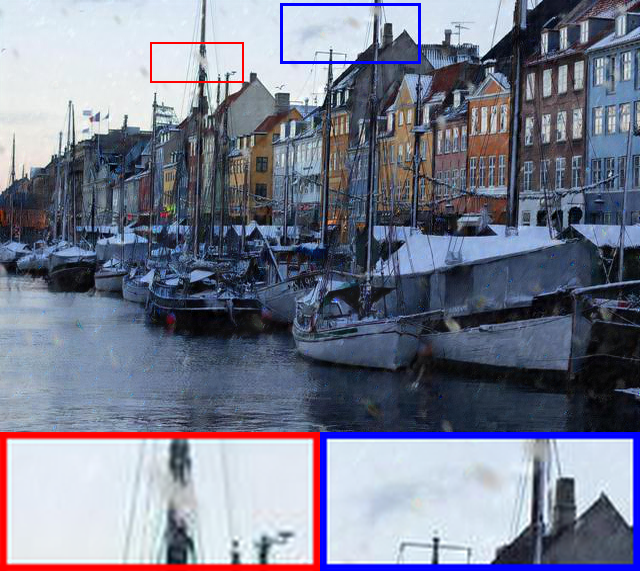} \\
\includegraphics[width=0.192\textwidth]{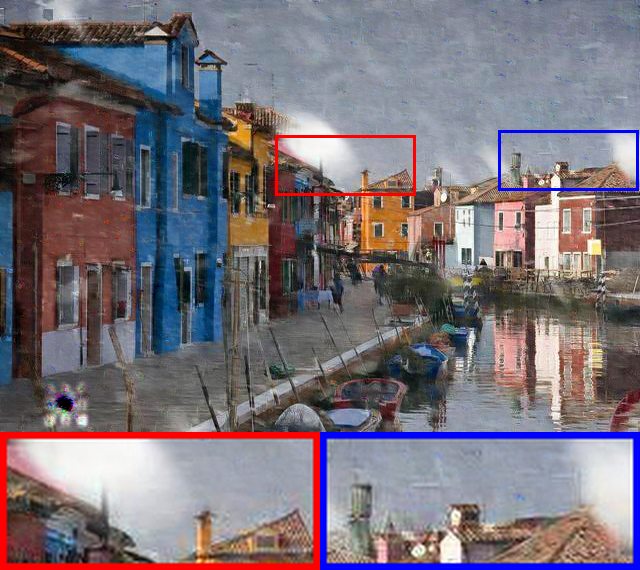} 
\end{tabular}%
}\hspace{-.45cm}
\subfloat[DDMSNet~\cite{zhang2021deep}\label{fig-snow:ddmsnet}]{%
\begin{tabular}{c}
\includegraphics[width=0.192\textwidth]{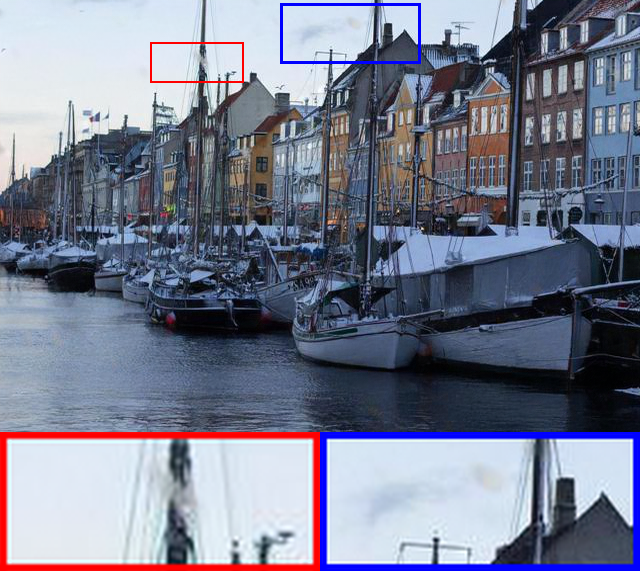} \\
\includegraphics[width=0.192\textwidth]{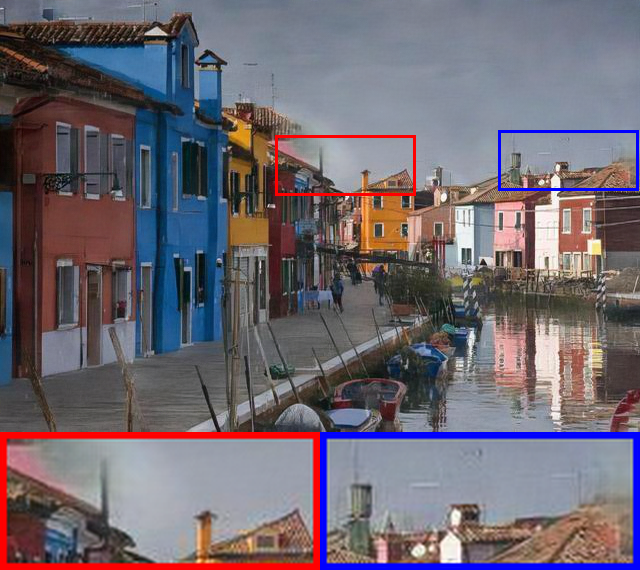}
\end{tabular}%
}\hspace{-.45cm}
\subfloat[\textbf{Ours (SnowDiff$_{64}$)}\label{fig-snow:ours}]{%
\begin{tabular}{c}
\includegraphics[width=0.192\textwidth]{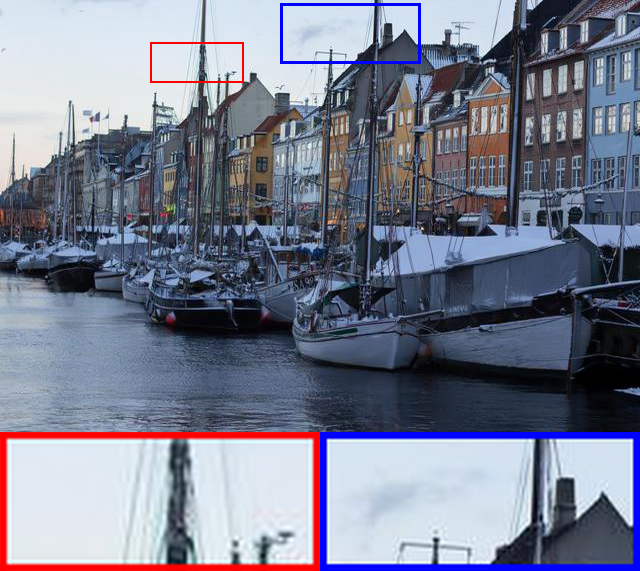} \\
\includegraphics[width=0.192\textwidth]{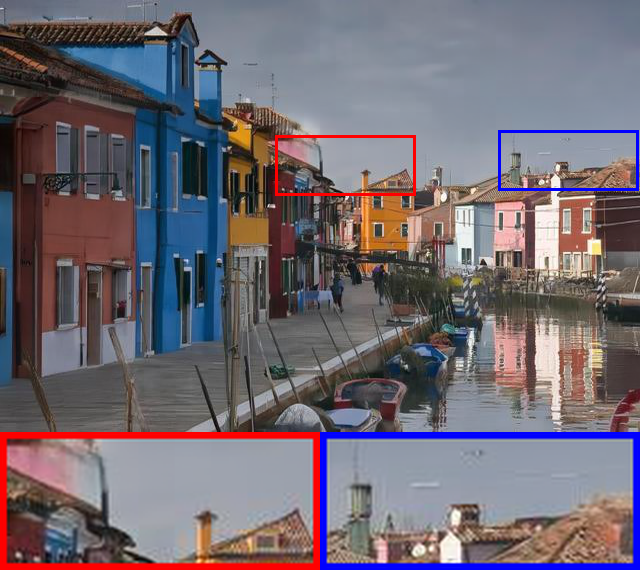}
\end{tabular}%
}\hspace{-.45cm}
\subfloat[Ground truth\label{fig-snow:gt}]{%
\begin{tabular}{c}
\includegraphics[width=0.192\textwidth]{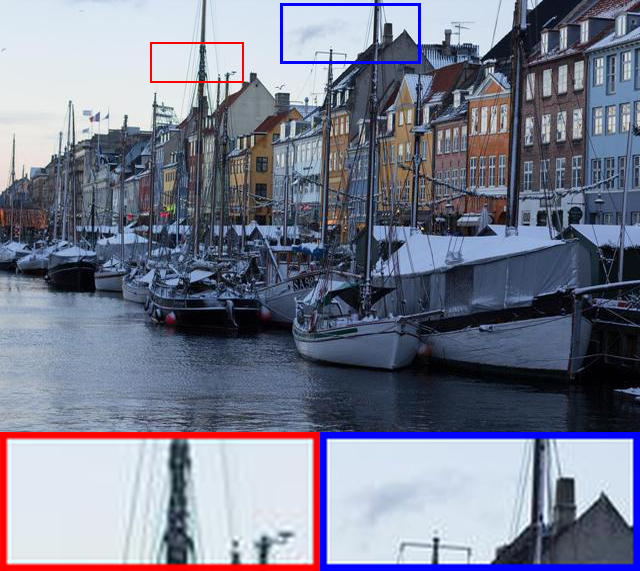} \\
\includegraphics[width=0.192\textwidth]{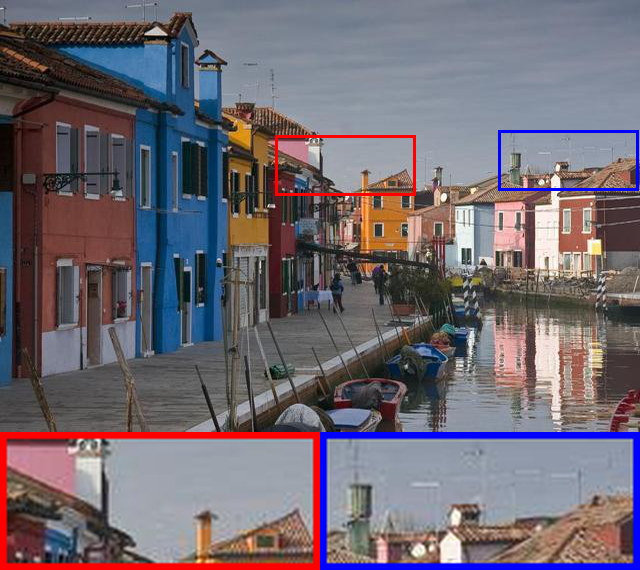}
\end{tabular}%
}\hspace{-.45cm}
\caption{Qualitative reconstruction comparisons of our best model on SnowTest100K test samples with DesnowNet~\cite{liu2018desnownet} and DDMSNet~\cite{zhang2021deep}.}
\label{fig:snow_reconstructions}
\end{figure*}

\subsection{Comparison Methods and Evaluation Metrics}
\label{sec:eval_metrics}

We perform comparisons of our weather-specific models with several state-of-the-art methods discussed in Section~\ref{sec:bg_restoration} for image desnowing~\cite{li2018recurrent,Wang:2019SPANet,chen2020jstasr,liu2018desnownet,zhang2021deep}, combined image deraining and dehazing~\cite{zhu2017unpaired,isola2017image,li2019heavy,Jiang:2021TIP,zamir2021multi}, and removing raindrops~\cite{isola2017image,liu2019dual,quan2019deep,qian2018attentive,xiao2022image}.
We compare WeatherDiff$_p$ with two state-of-the-art multi-weather image restoration methods: All-in-One~\cite{Li:2020CVPR}, which utilizes a multi-encoder and decoder pipeline with a neural architecture search mechanism,
and TransWeather~\cite{Valanarasu:2022CVPR}, which exploits an end-to-end vision transformer.
Notably, both of these works were presented for multi-weather image restoration using the same three benchmark datasets.

Our comparison method choices were mainly driven in accordance with the baselines from~\cite{Valanarasu:2022CVPR,Li:2020CVPR}, as well as methods that are in directly comparable setting since they either reported identical test set evaluations with the datasets we used, or publicly provided their pretrained models.

Quantitative evaluations between ground truth and restored images were performed via the conventional peak signal-to-noise ratio (PSNR)~\cite{huynh2008scope} and structural similarity (SSIM)~\cite{wang2004image} metrics.
We evaluated PSNR and SSIM based on the luminance channel Y of the YCbCr color space in accordance with the previous convention \cite{qian2018attentive,Valanarasu:2022CVPR,zamir2021multi,xiao2022image}.
We also used two other metrics for reference-free quality assessment of real-world restoration performance, namely the Naturalness Image Quality Evaluator (NIQE)~\cite{mittal2012making}, and Integrated Local NIQE (IL-NIQE)~\cite{zhang2015feature} scores. Better perceptual image quality leads to lower NIQE and IL-NIQE scores.

\subsection{Weather-Specific Image Restoration Results}
\label{sec:weatherspecific}

Figure~\ref{tab:image_restoration} presents our quantitative evaluations.
The top half of the tables contain results from weather-specific image restoration, where we show $S=10$ sampling time steps for $p=64$, and $S=50$ for $p=128$ (see Section~1.2 of Supplementary Materials for other choices, where sometimes better results can be achieved by tuning $S$ for each task individually).
Our models achieve performances superior to all compared existing methods on all tasks.
For image desnowing and combined deraining and dehazing tasks, our 64x64 patch models yield the best results (i.e., 36.59/0.9626 on Snow100K-S, 30.43/0.9145 on Snow100K-L and 28.38/0.9320 on Outdoor-Rain).
For removing raindrops, with both input patch resolutions we outperform the recent image de-raining transformers~\cite{xiao2022image} with RainDropDiff$_{128}$ having the best PSNR of 32.52.

\begin{figure*}[!ht]
\subfloat[Input\label{fig-rainfog:input}]{%
\begin{tabular}{c}
\includegraphics[width=0.192\textwidth]{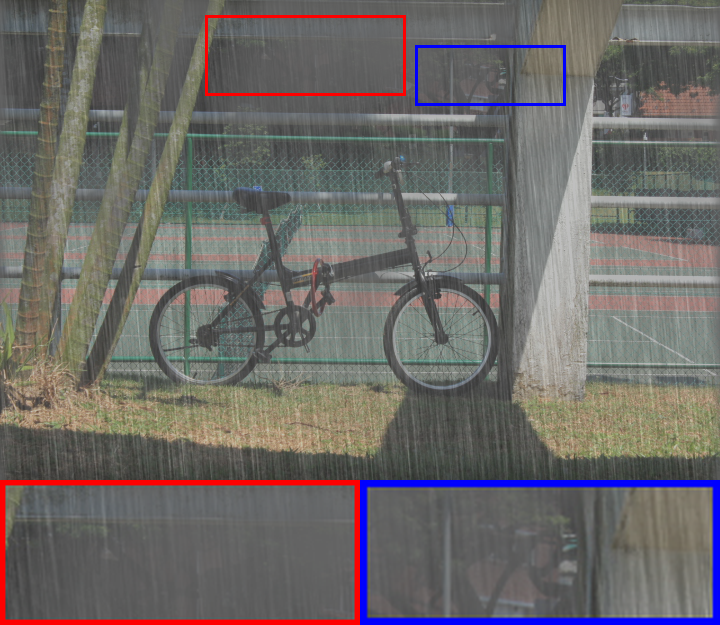} \\
\includegraphics[width=0.192\textwidth]{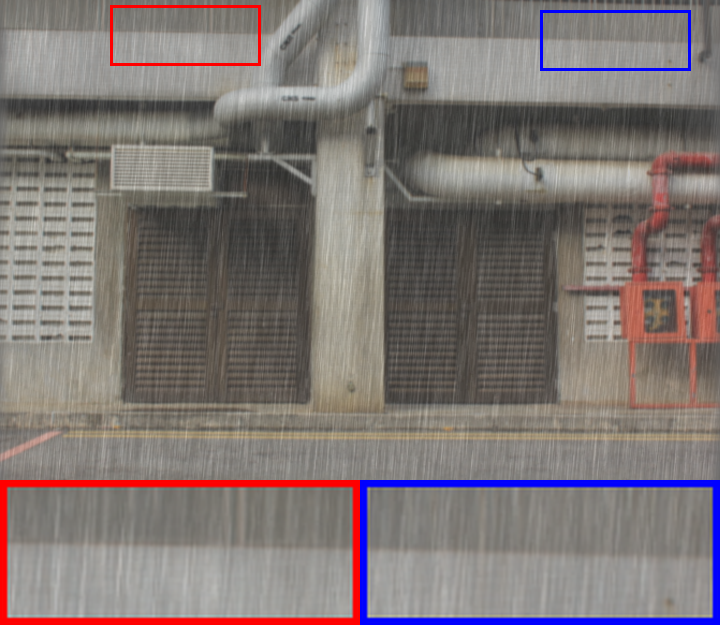}
\end{tabular}%
}\hspace{-.45cm}
\subfloat[HRGAN~\cite{li2019heavy}\label{fig-rainfog:hrgan}]{%
\begin{tabular}{c}
\includegraphics[width=0.192\textwidth]{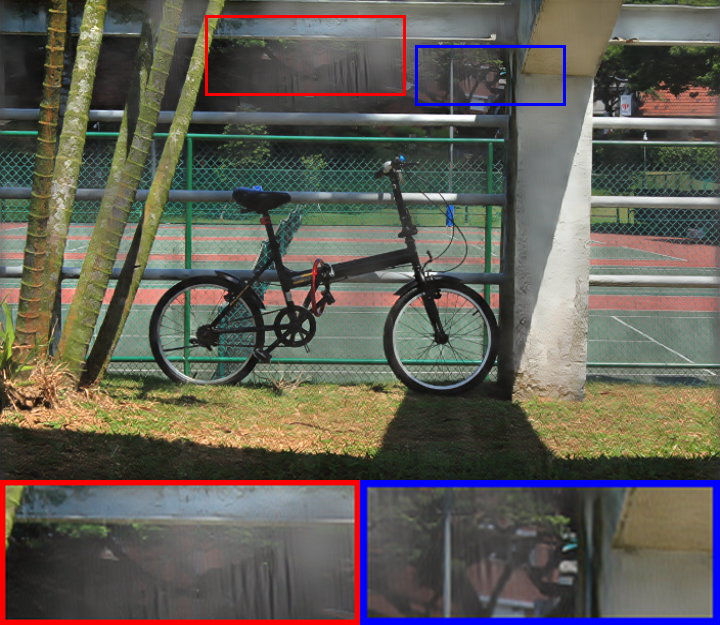} \\
\includegraphics[width=0.192\textwidth]{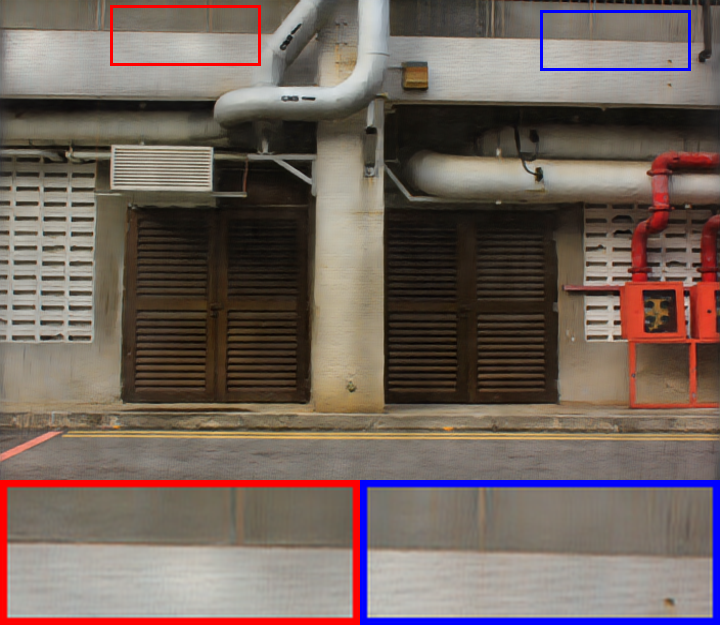}
\end{tabular}%
}\hspace{-.45cm}
\subfloat[MPRNet~\cite{zamir2021multi}\label{fig-rainfog:mprnet}]{%
\begin{tabular}{c}
\includegraphics[width=0.192\textwidth]{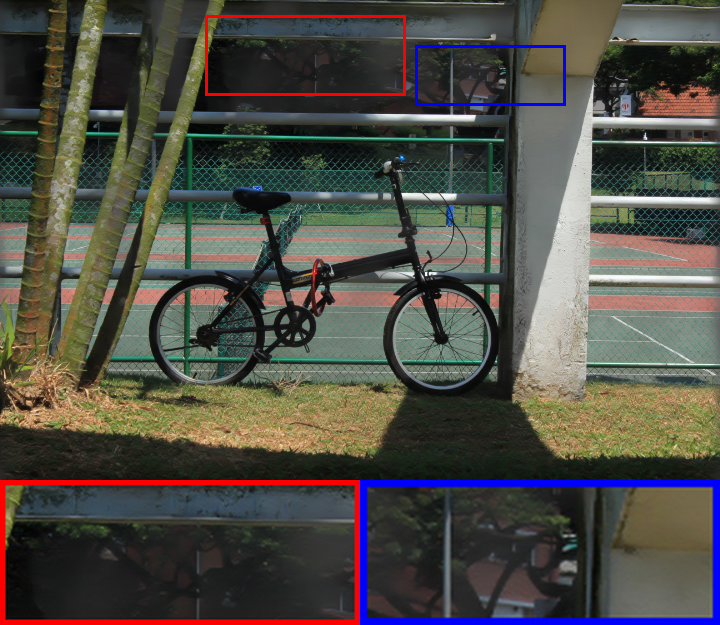} \\
\includegraphics[width=0.192\textwidth]{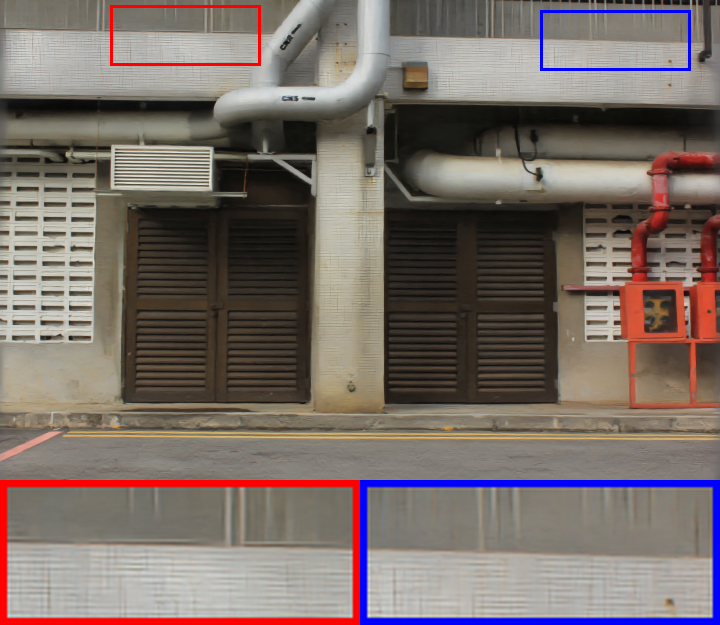}
\end{tabular}%
}\hspace{-.45cm}
\subfloat[\textbf{Ours (RainHazeDiff$_{64}$)}\label{fig-rainfog:ours}]{%
\begin{tabular}{c}
\includegraphics[width=0.192\textwidth]{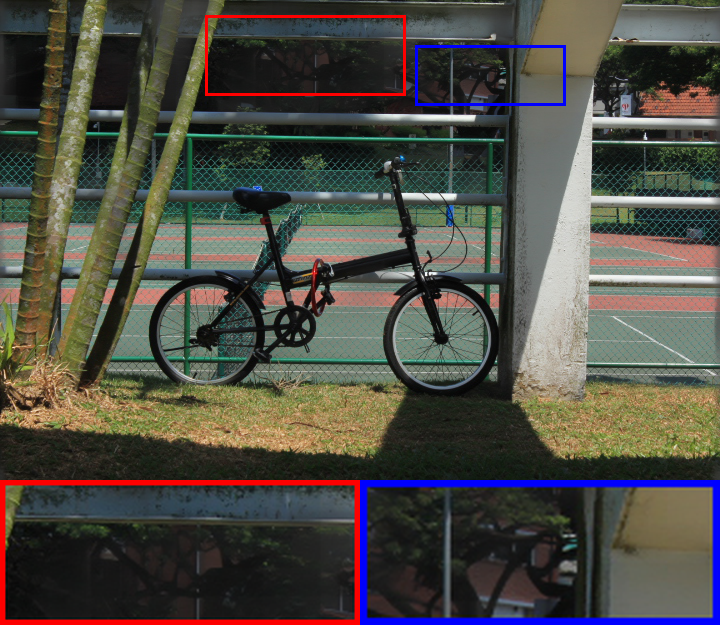} \\
\includegraphics[width=0.192\textwidth]{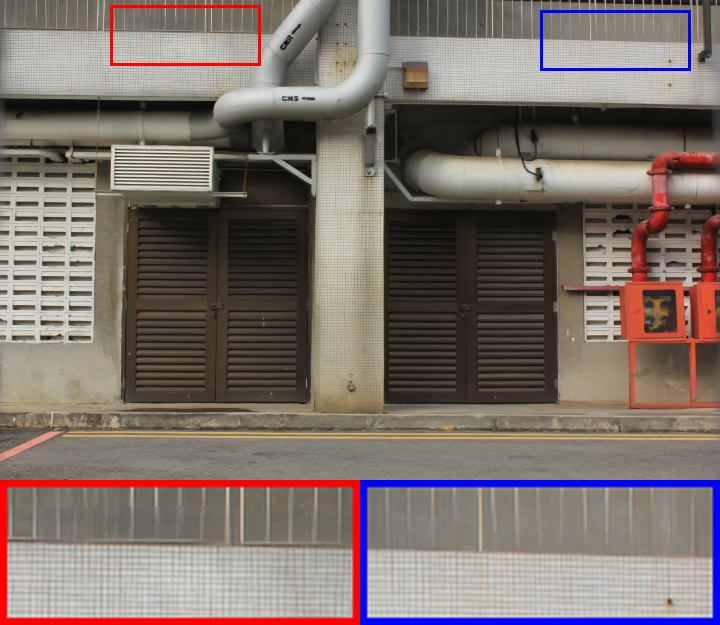}
\end{tabular}%
}\hspace{-.45cm}
\subfloat[Ground truth\label{fig-rainfog:gt}]{%
\begin{tabular}{c}
\includegraphics[width=0.192\textwidth]{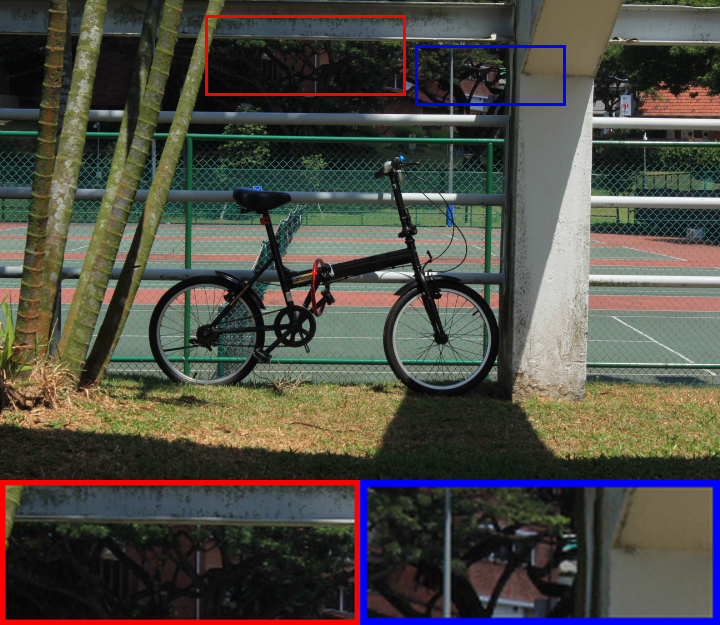} \\
\includegraphics[width=0.192\textwidth]{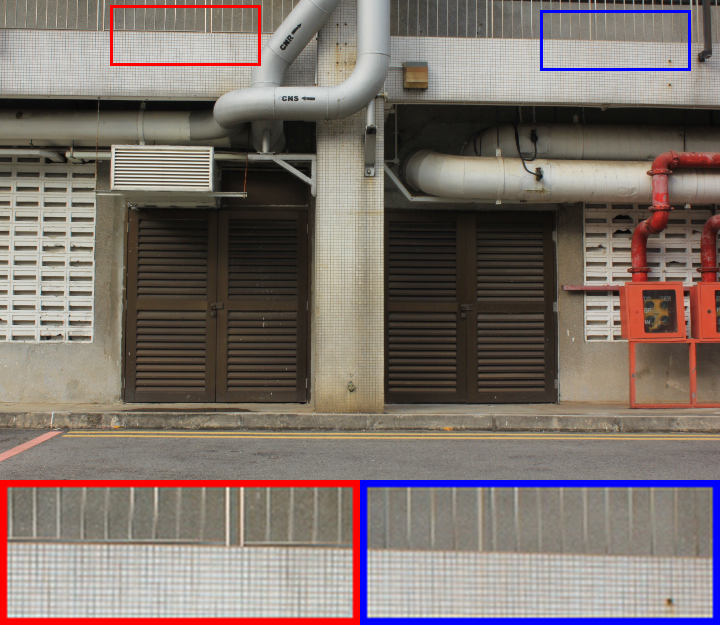}
\end{tabular}%
}\hspace{-.45cm}
\caption{Qualitative reconstruction comparisons of our best model on Outdoor-Rain test samples with HRGAN~\cite{li2019heavy} and MPRNet~\cite{liu2018desnownet}.}
\label{fig:rainhaze_reconstructions}
\end{figure*}

\begin{figure*}[!t]
\subfloat[Input\label{fig-raindrop:input}]{%
\begin{tabular}{c}
\includegraphics[width=0.192\textwidth]{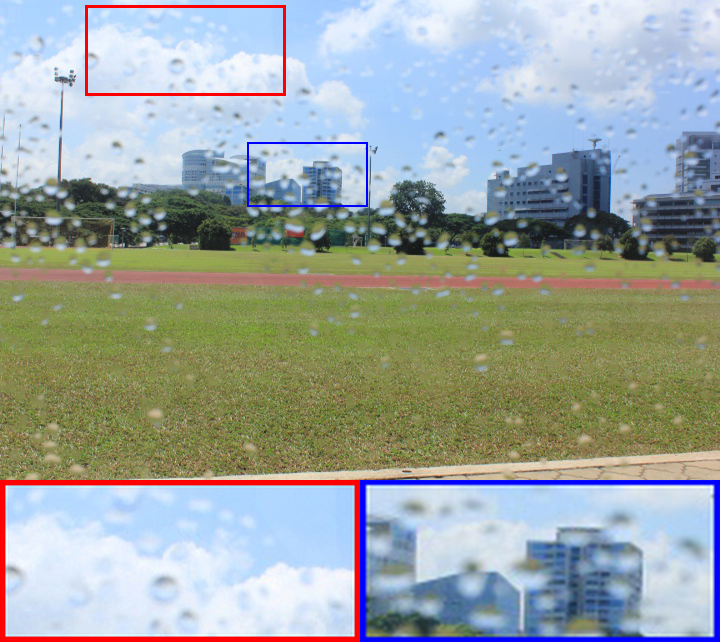} \\
\includegraphics[width=0.192\textwidth]{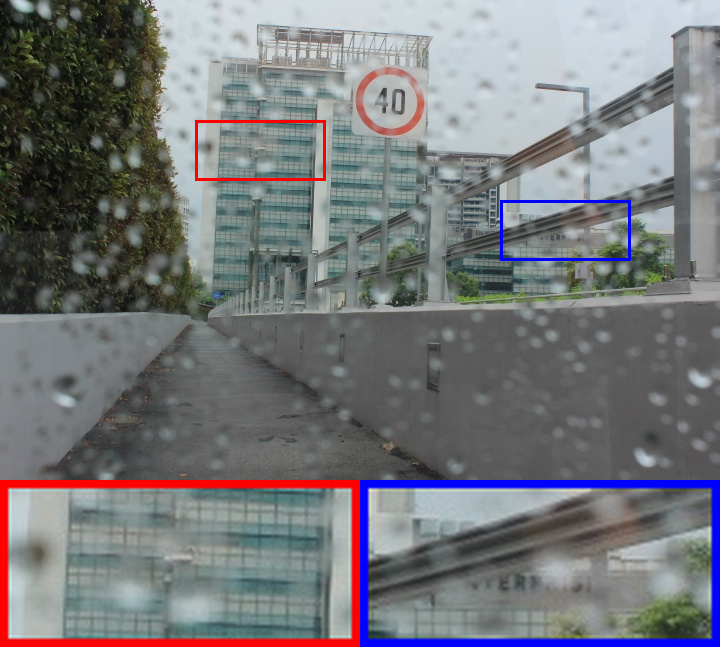}
\end{tabular}%
}\hspace{-.45cm}
\subfloat[RaindropAttn~\cite{quan2019deep}\label{fig-raindrop:raindropattn}]{%
\begin{tabular}{c}
\includegraphics[width=0.192\textwidth]{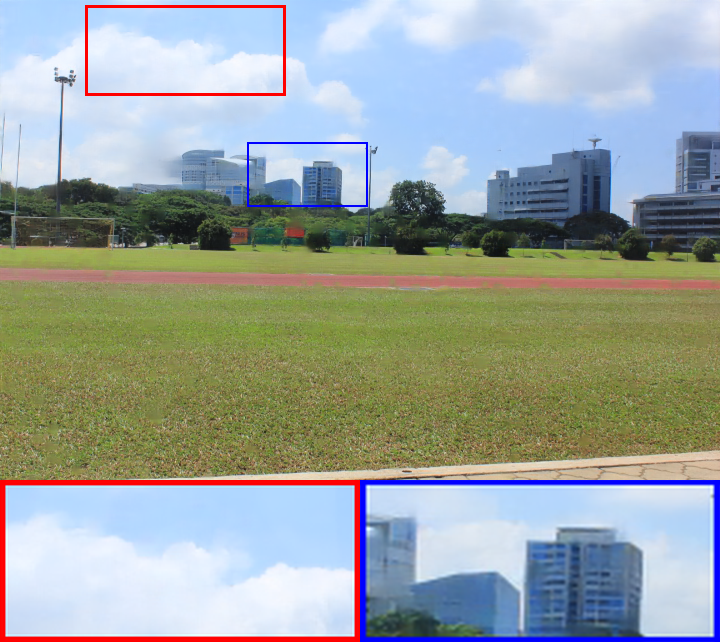} \\
\includegraphics[width=0.192\textwidth]{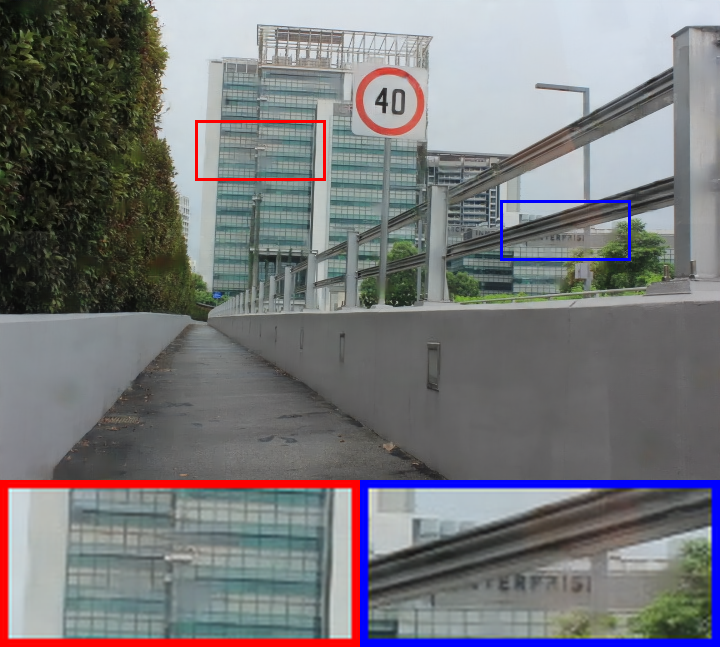}
\end{tabular}%
}\hspace{-.45cm}
\subfloat[AttentiveGAN~\cite{qian2018attentive}\label{fig-raindrop:attentgan}]{%
\begin{tabular}{c}
\includegraphics[width=0.192\textwidth]{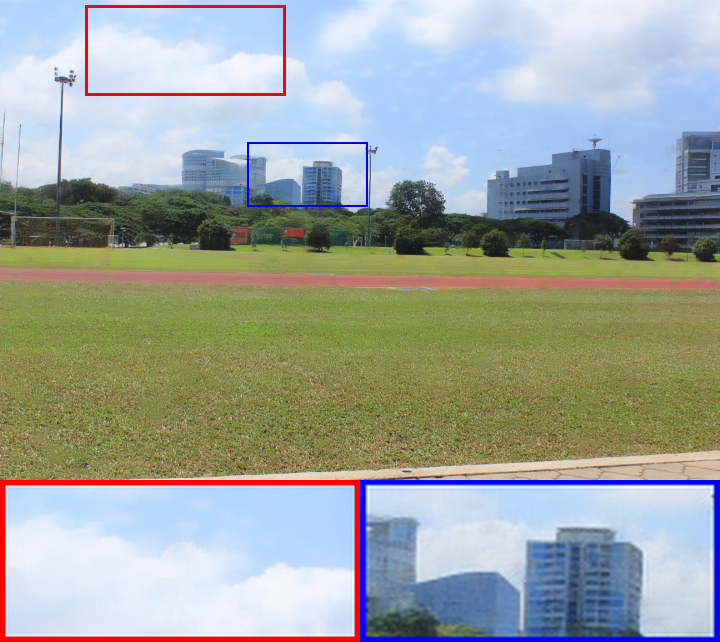} \\
\includegraphics[width=0.192\textwidth]{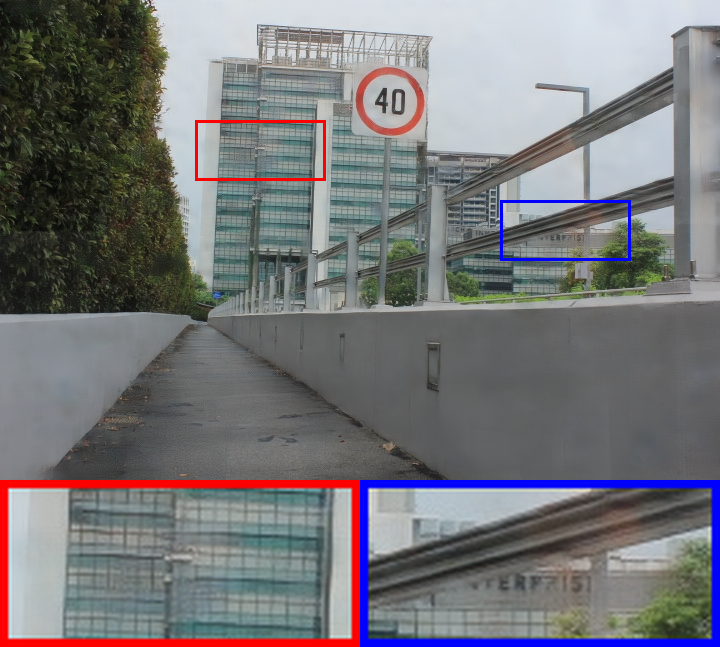} 
\end{tabular}%
}\hspace{-.45cm}
\subfloat[\textbf{Ours (RainDropDiff$_{128}$)}\label{fig-raindrop:ours}]{%
\begin{tabular}{c}
\includegraphics[width=0.192\textwidth]{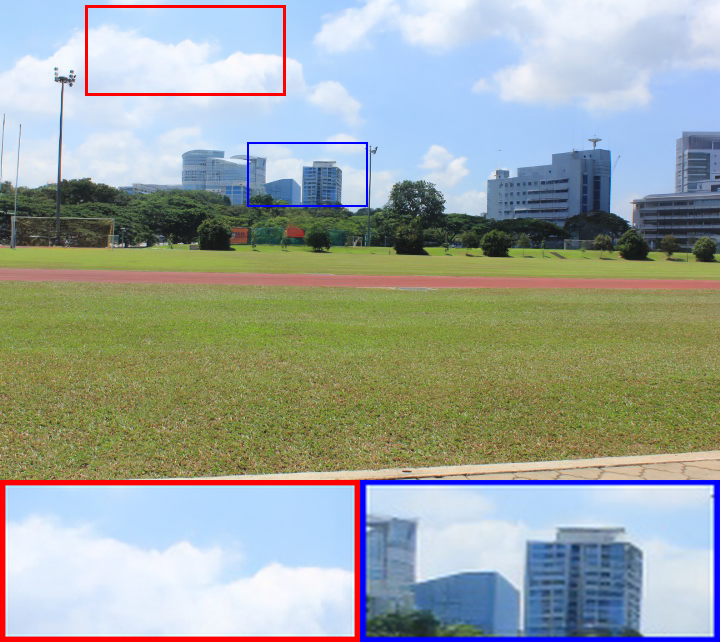} \\
\includegraphics[width=0.192\textwidth]{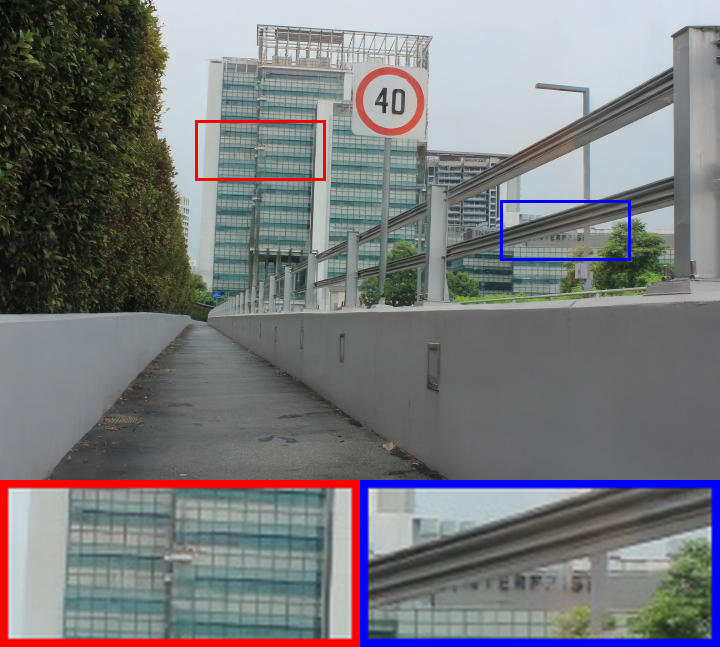}
\end{tabular}%
}\hspace{-.45cm}
\subfloat[Ground truth\label{fig-raindrop:gt}]{%
\begin{tabular}{c}
\includegraphics[width=0.192\textwidth]{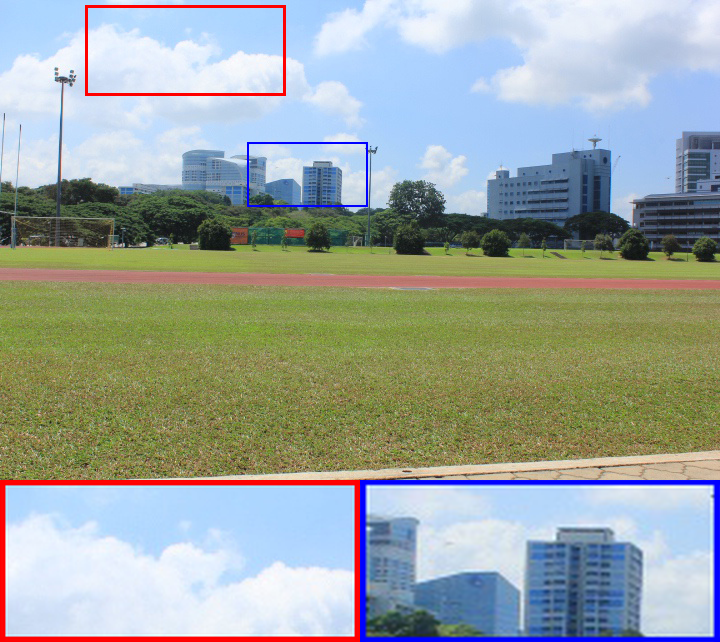} \\
\includegraphics[width=0.192\textwidth]{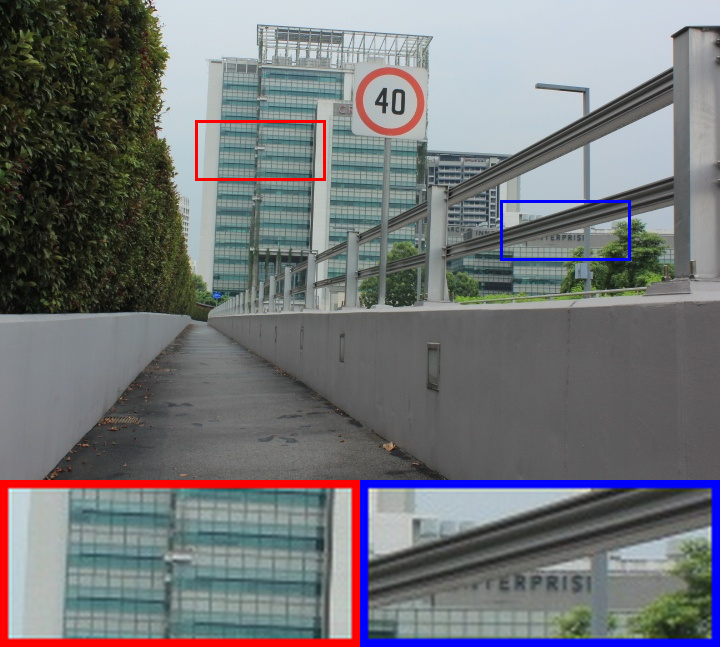}
\end{tabular}%
}\hspace{-.45cm}
\caption{Qualitative reconstruction comparisons of our best model on Raindrop test samples with RaindropAttn~\cite{quan2019deep} and AttentiveGAN~\cite{qian2018attentive}.}
\label{fig:raindrop_reconstructions}
\end{figure*}

Figure~\ref{fig:snow_reconstructions} depicts some visualizations of image desnowing reconstructions for sample test images, comparing our method with DesnowNet~\cite{liu2018desnownet} and DDMSNet~\cite{zhang2021deep}.
As illustrated, while DDMSNet appears to achieve noticable higher visual quality than DeSnowNet in reconstructions, our method SnowDiff$_{64}$ shows remarkable restoration quality in fine details (enlarged in red and blue bounding boxes).

Figure~\ref{fig:rainhaze_reconstructions} depicts visualizations on sample Outdoor-Rain test images, demonstrating the superiority of our model RainHazeDiff$_{64}$ over HRGAN~\cite{li2019heavy} and MPRNet~\cite{quan2019deep}.
In particular, smoothing effects from dehazing results in loss of details in reconstructions with other methods, while our model can recover these (e.g., second example in Figure~\ref{fig:rainhaze_reconstructions}, metal railing lines enlarged in the bounding boxes).

Figure~\ref{fig:raindrop_reconstructions} visualizes raindrop removal examples, comparing our best model RainDropDiff$_{128}$ with AttentiveGAN~\cite{qian2018attentive} and RaindropAttn~\cite{quan2019deep}.
Note that we particularly illustrate HRGAN on Outdoor-Rain, and AttentiveGAN on RainDrop test sets, since these approaches are earlier generative modeling based applications to this problem with GANs.
Our models generate more resembling reconstructions to the ground truths in all comparisons, and diffusion generative modeling significantly outperforms the ones based on GANs.
We could not present visual comparisons to IDT~\cite{xiao2022image} due to publicly unavailable implementations.

\subsection{Multi-Weather Image Restoration Results}
\label{sec:multiweather}

The bottom half of Figure~\ref{tab:image_restoration} presents quantitative evaluations for multi-weather image restoration in comparison to All-in-One and TransWeather, where we show $S=25$ for $p=64$, and $S=50$ for $p=128$ (see Section~1.2 of Supplementary Materials for other choices, where sometimes better results can be achieved by tuning $S$ for each task individually).
We present PSNR/SSIM for publicly available TransWeather predictions with our definitions from Section~\ref{sec:eval_metrics}, which gave different results than reported in~\cite{Valanarasu:2022CVPR}.

\begin{figure*}[!ht]
\subfloat[Input\label{fig-realsnow:input}]{%
\begin{tabular}{c}
\includegraphics[width=0.323\textwidth]{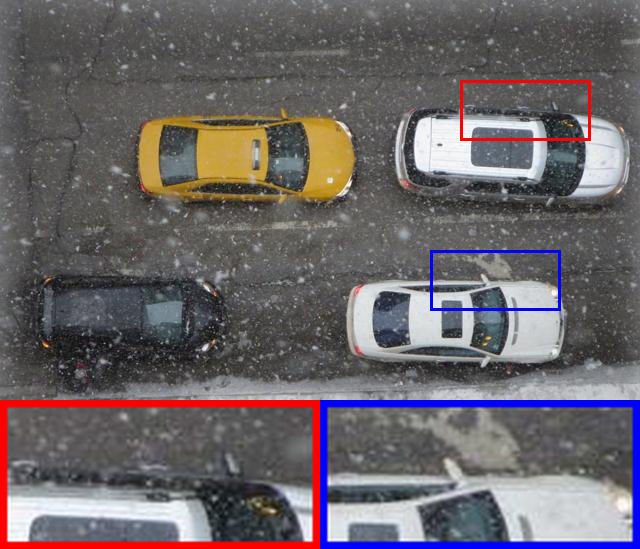} \\
\includegraphics[width=0.323\textwidth]{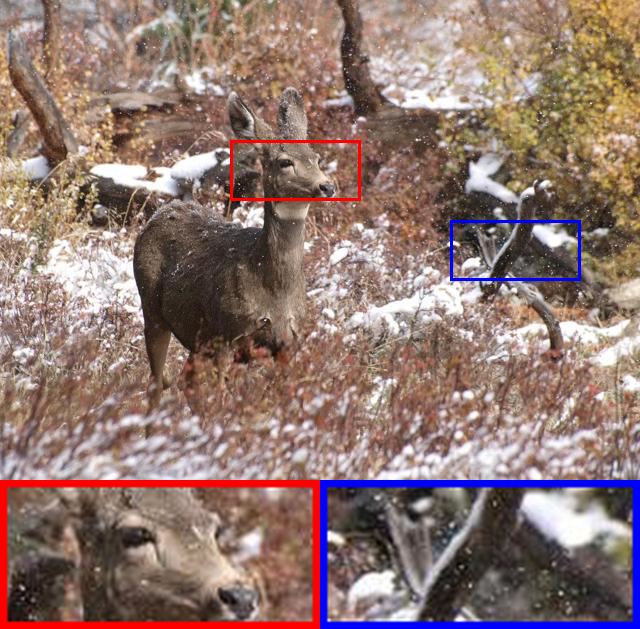}
\end{tabular}%
}\hspace{-.45cm}
\subfloat[TransWeather~\cite{Valanarasu:2022CVPR}\label{fig-realsnow:transweather}]{%
\begin{tabular}{c}
\includegraphics[width=0.323\textwidth]{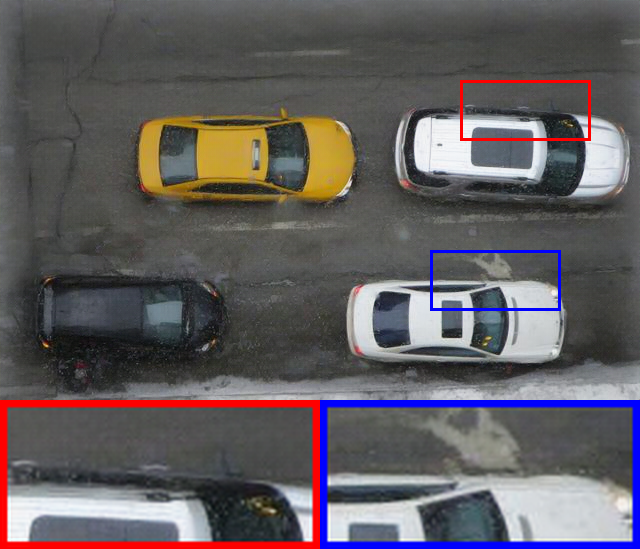} \\
\includegraphics[width=0.323\textwidth]{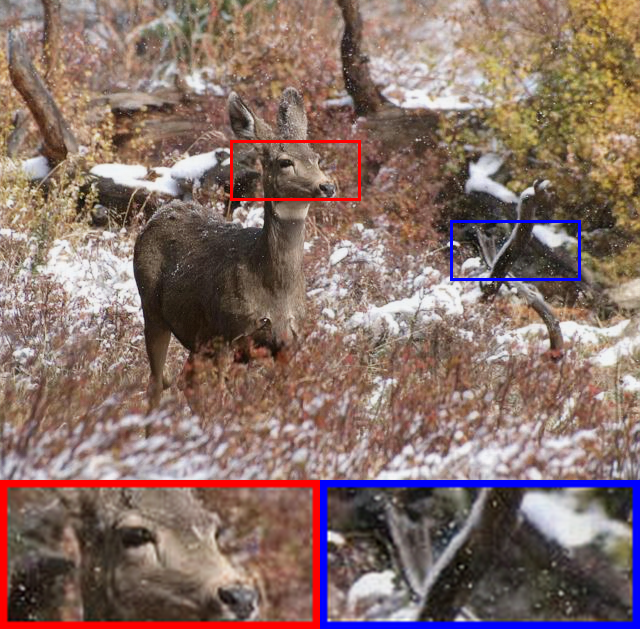}
\end{tabular}%
}\hspace{-.45cm}
\subfloat[\textbf{Ours (WeatherDiff$_{64}$)}\label{fig-realsnow:ours}]{%
\begin{tabular}{c}
\includegraphics[width=0.323\textwidth]{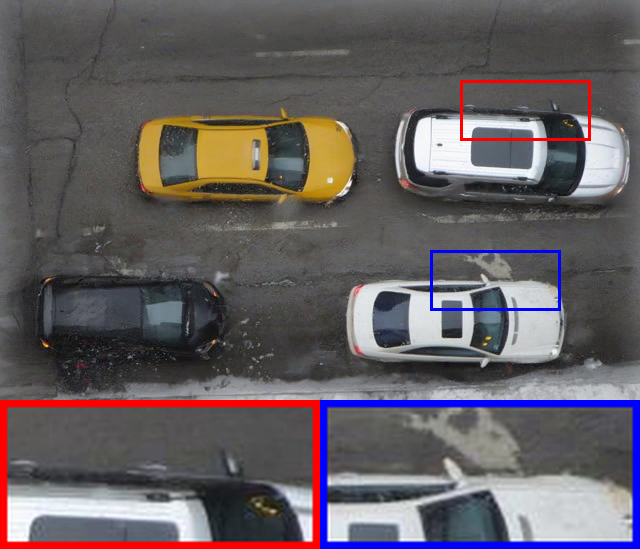} \\
\includegraphics[width=0.323\textwidth]{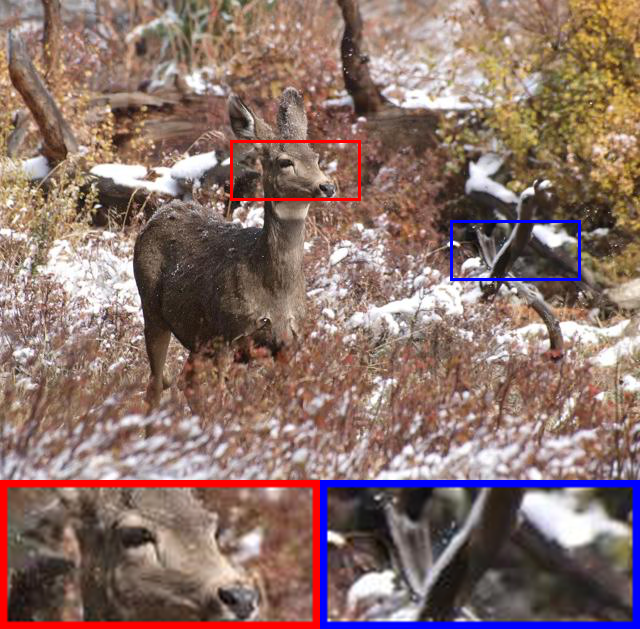}
\end{tabular}%
}\hspace{-.45cm}
\caption{Comparison of real-world snowy image restoration examples using TransWeather~\cite{Valanarasu:2022CVPR} and WeatherDiff$_{64}$. In the above example TransWeather mistakenly removes the side view mirrors from both cars, however yields cleaner restorations than our method around the black car. In the below example our method obtains better removal of tiny snowflakes from images when viewed in detail.}
\label{fig:realistic_snow_reconstructions}
\end{figure*}

Generally our method yields exceptional image quality and ground truth similarity on all three test sets. For the image desnowing task, WeatherDiff$_{64}$ achieves the best PSNR/SSIM metrics with 35.83/0.9566 and 30.09/0.9041 for Snow100K-S and Snow100K-L respectively.
Notably, on the combined image deraining and dehazing task, WeatherDiff$_{64}$ and WeatherDiff$_{128}$ yields better PSNR values of 29.64 and 29.72 respectively, which also outperforms all dedicated weather-specific models at the above half of Figure~\ref{tab:outdoorrain}.
This is particularly important as WeatherDiff$_{p}$ significantly outperforms our RainHazeDiff$_{p}$ models on this task. This indicates an improvement of the background generative capability when combined with other tasks and datasets.
None of the existing multi-weather restoration methods showed a similar knowledge transfer in comparison to their weather-specific counterparts. 

Our models are only outperformed in a single metric, by All-in-One for PSNR on the RainDrop task (All-in-One: 31.12, Ours: 30.71).
Nevertheless our results show better ground truth similarity for this case (All-in-One: 0.9268, Ours: 0.9312).
These results demonstrate that WeatherDiff models can successfully learn the underlying data distribution under several adverse weather corruption tasks.

\begin{figure*}[!ht]
\subfloat[Input\label{fig-realrainds:input}]{%
\begin{tabular}{c}
\includegraphics[width=0.323\textwidth]{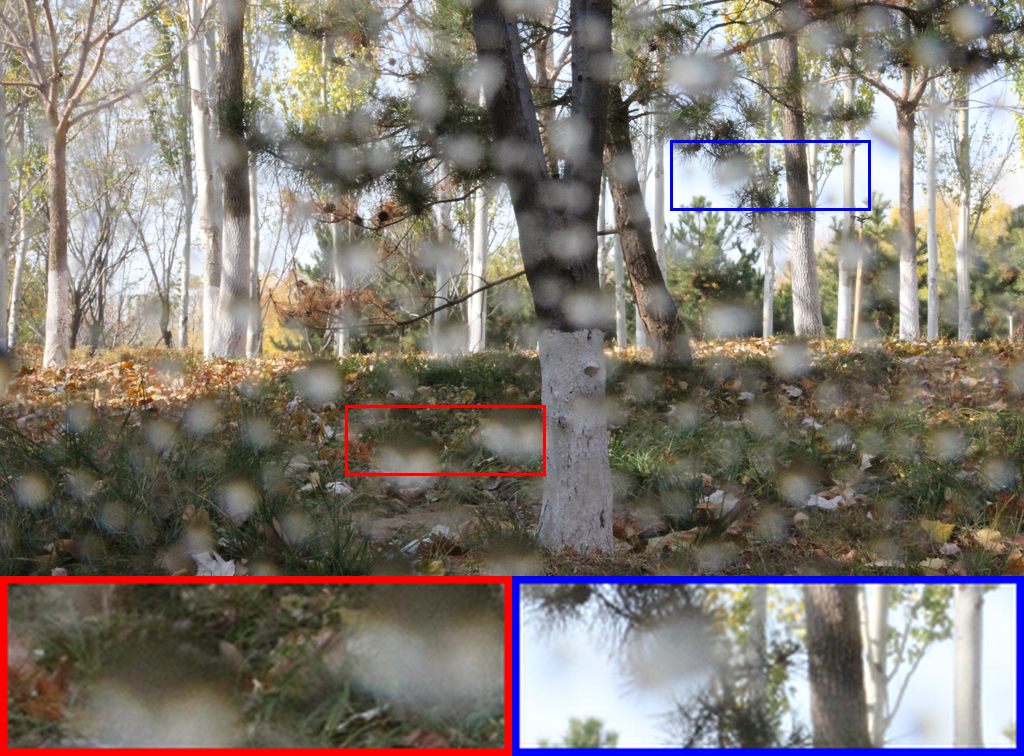} \\
\includegraphics[width=0.323\textwidth]{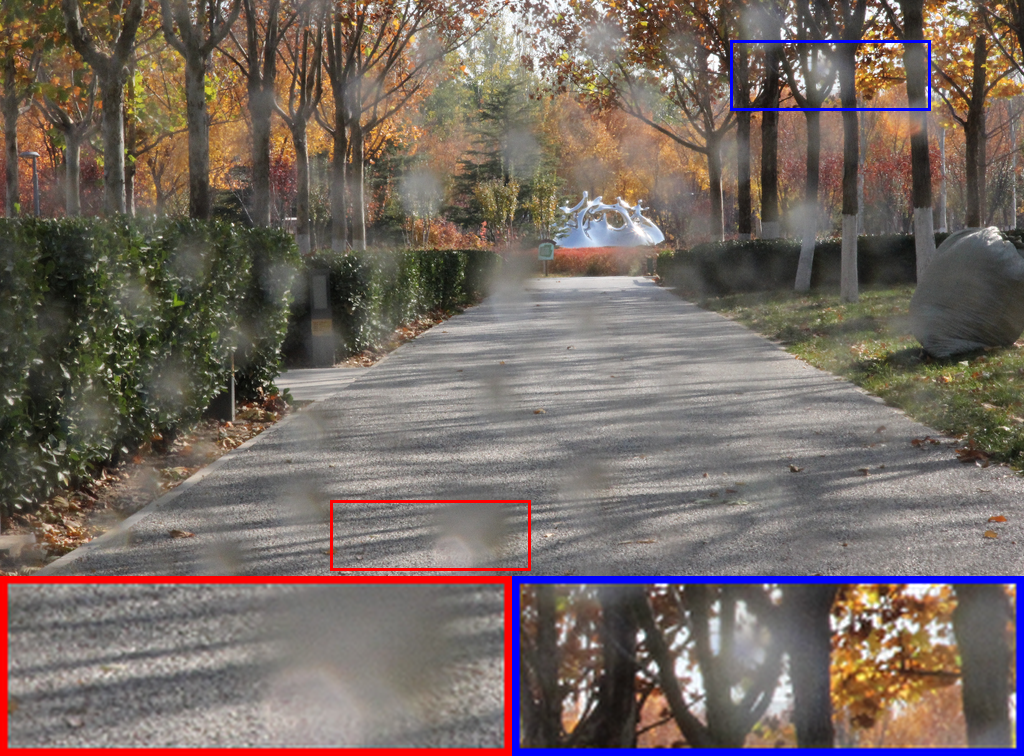}
\end{tabular}%
}\hspace{-.45cm}
\subfloat[TransWeather~\cite{Valanarasu:2022CVPR}\label{fig-realrainds:transweather}]{%
\begin{tabular}{c}
\includegraphics[width=0.323\textwidth]{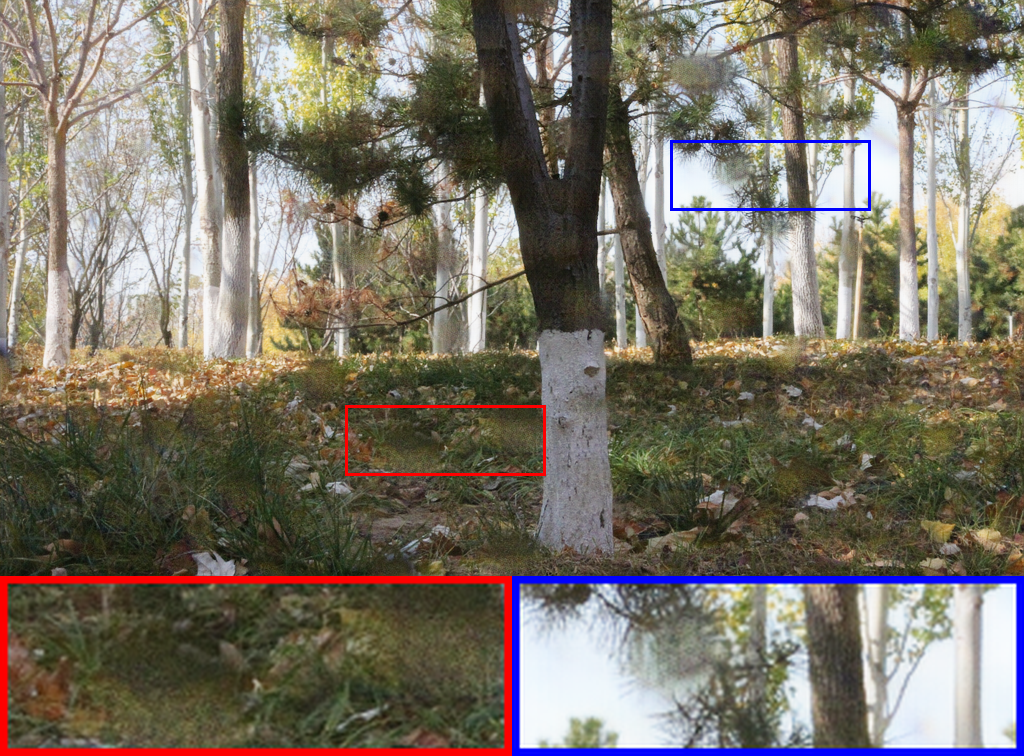} \\
\includegraphics[width=0.323\textwidth]{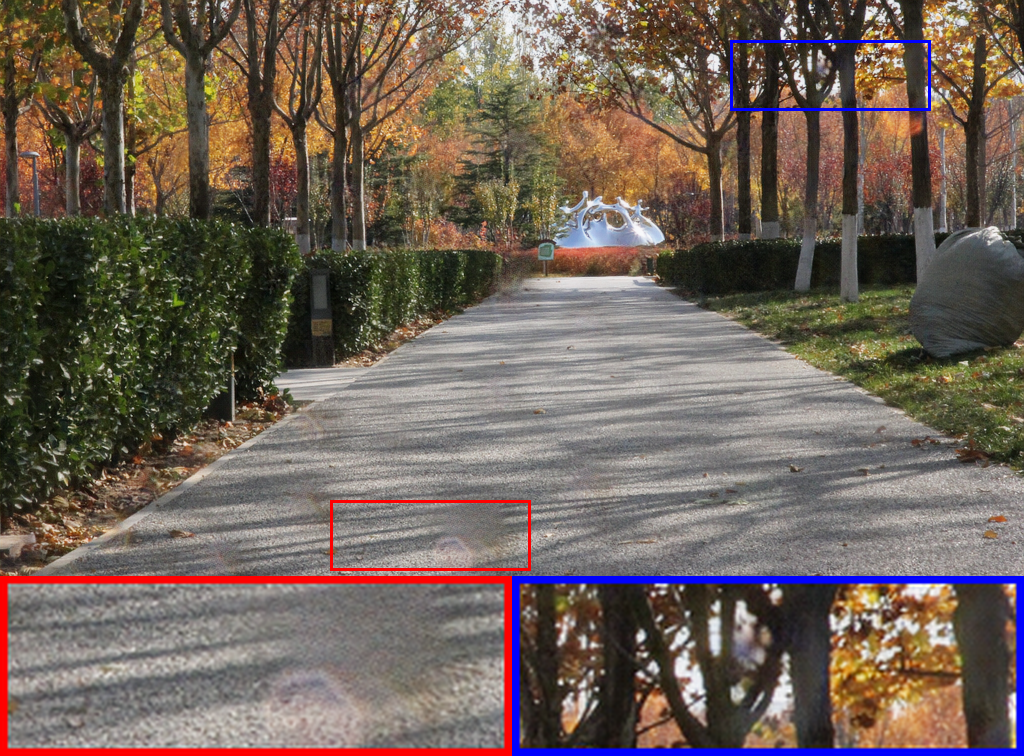}
\end{tabular}%
}\hspace{-.45cm}
\subfloat[\textbf{Ours (WeatherDiff$_{64}$)}\label{fig-realrainds:ours}]{%
\begin{tabular}{c}
\includegraphics[width=0.323\textwidth]{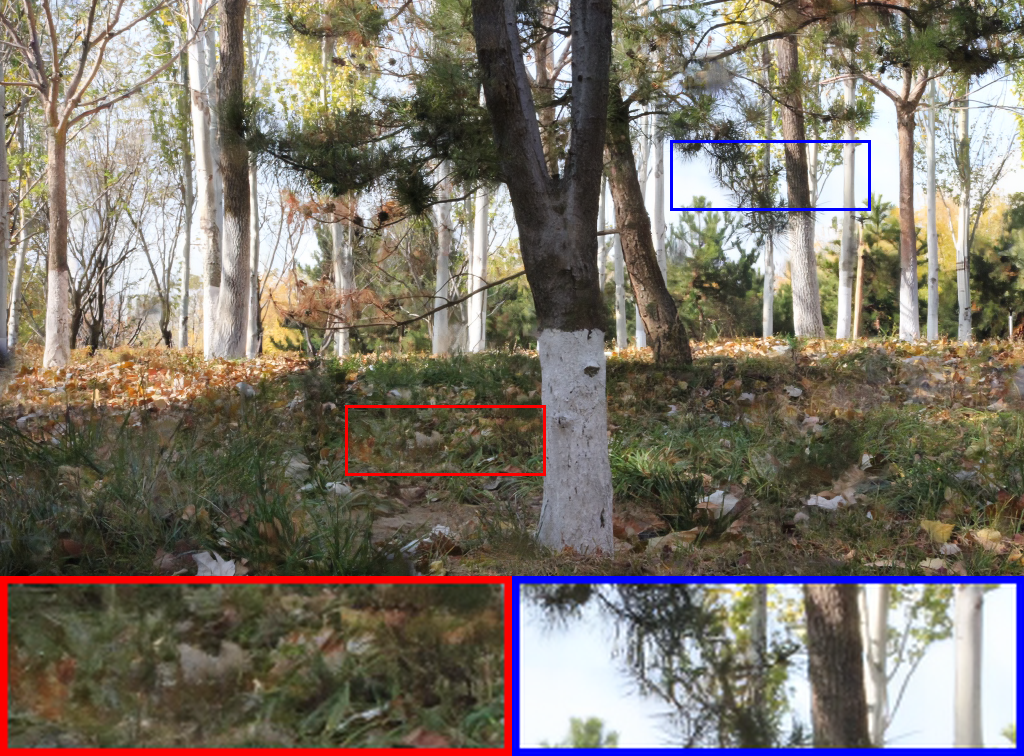} \\
\includegraphics[width=0.323\textwidth]{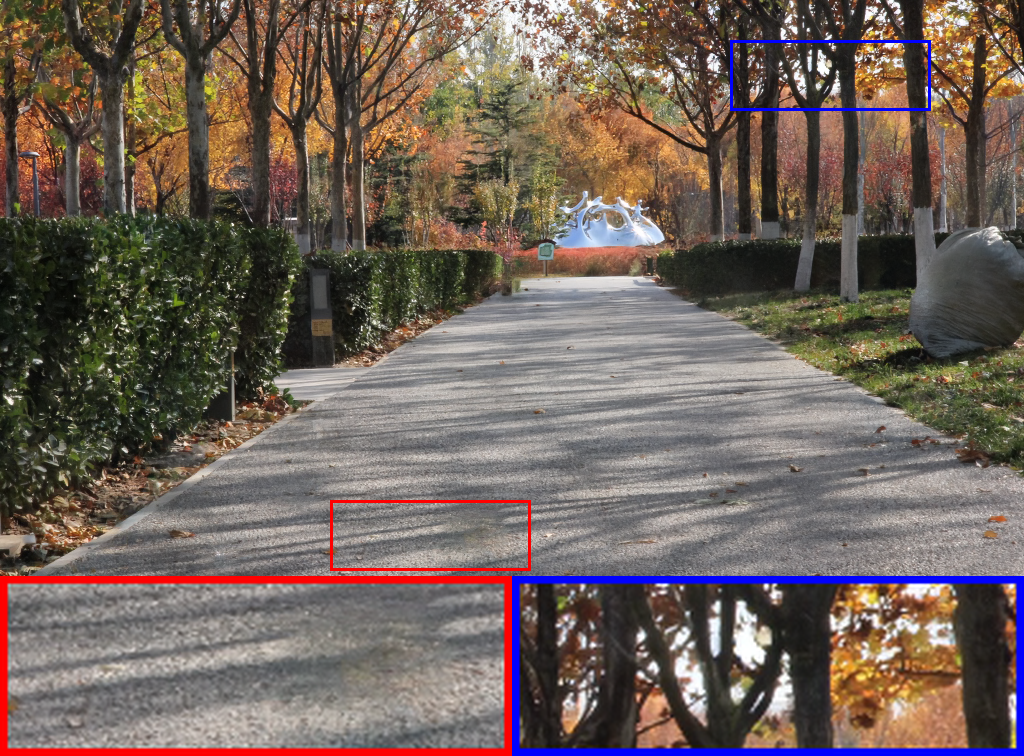}
\end{tabular}%
}\hspace{-.45cm}
\caption{Comparison of real-world raindrop image restoration examples using TransWeather~\cite{Valanarasu:2022CVPR} and WeatherDiff$_{64}$. Our method generates creative reconstructions in the above example with stones on the grass and sharper leaves on branches, whereas TransWeather smoothes out many details. In the below example, very bright raindrop artifacts could not be restored by TransWeather while our model recovers these.}
\label{fig:realistic_rainds_reconstructions}
\end{figure*}

\subsection{Weather Restoration Generalization from Synthetic to Real-World Images}
\label{sec:real_restoration}

We evaluate our models trained on synthetic data with real-world image restoration test cases. 
For visual illustrations we compare our best performing WeatherDiff$_{64}$ model with the recent TransWeather network, which are both specialized on multi-weather restoration.
Figure~\ref{fig:realistic_snow_reconstructions} presents qualitative image desnowing comparisons for selected images with light snow from the miscellaneous realistic snowy images set Snow100K-Real~\cite{liu2018desnownet}.
First example in Figure~\ref{fig:realistic_snow_reconstructions} shows a case where reconstructions by TransWeather removes the side view mirrors of cars, whereas our model preserves this detail (enlarged in the bounding boxes).
On the other hand, TransWeather gave cleaner restorations than our method around the black car overall.
In the second example, clearer reconstructions with our model can be observed for a detailed image with light snow artifacts.

We also included additional real-world restoration test cases from the raindrop removal test set of the RainDS dataset presented by~\cite{Quan:2021CVPR} which consisted of 97 real test images.
Figure~\ref{fig:realistic_rainds_reconstructions} presents qualitative comparisons for removing raindrops from real images using the same multi-weather restoration models.
First example in Figure~\ref{fig:realistic_rainds_reconstructions} depicts a detailed image where TransWeather reconstructions removes partly obstructed background components (i.e., leaves and stones), whereas our generative model completes these details during restoration.
Second example shows a case with very bright raindrop artifacts on the camera sensor which were not completely restored by TransWeather, whereas our model is comparably better.
We provide more visual examples in Section~2 of Supplementary Materials.

\begin{table}[t!]
\caption{Quantitative NIQE and IL-NIQE score comparisons on real-world image datasets with multi-weather restoration models. Best and second best values are indicated with bold text and underlined text respectively.}
\label{tab:realworld_niqe}
\scalebox{0.98}{
\begin{tabular}{p{2.3cm} c c c c c c c c c}
\toprule
& \multicolumn{2}{c}{Snow100K-Real~\cite{liu2018desnownet}} &  \multicolumn{2}{c}{RainDS~\cite{Quan:2021CVPR}} \\
\cmidrule(l{.5em}r{.5em}){2-3}\cmidrule(l{.5em}r{.5em}){4-5}
& NIQE $\downarrow$ & IL-NIQE $\downarrow$ & NIQE $\downarrow$ & IL-NIQE $\downarrow$ \\
\midrule
TransWeather~\cite{Valanarasu:2022CVPR} & 3.161 & 22.207 & 4.005 & 22.512 \\
\midrule
\textbf{WeatherDiff$_{64}$} & \underline{2.985} & \underline{22.121} & \textbf{3.050} & \textbf{19.800} \\
\textbf{WeatherDiff$_{128}$} & \textbf{2.964} & \textbf{21.976} & \underline{3.642} & \underline{19.972} \\
\bottomrule
\end{tabular}}
\end{table}

Finally in Table~\ref{tab:realworld_niqe} we present our quantitative comparisons on these two real-world image restoration test sets based on reference-free image quality metrics.
Results show that our WeatherDiff$_{64}$ ($S=25$) and WeatherDiff$_{128}$ ($S=50$) models yield better perceptual image quality scores on both test sets, and significantly outperforms the state-of-the-art multi-weather restoration model TransWeather~\cite{Valanarasu:2022CVPR}.

\section{Discussion}
\label{sec:discussion}

We present a novel patch-based image restoration approach based on conditional denoising diffusion probabilistic models, to improve vision under adverse weather conditions. 
Our solution is shown to yield state-of-the-art performance on weather-specific and multi-weather image restoration tasks on benchmark datasets.
Notably, our method is general to any conditional diffusive generative modeling task with arbitrary sized images.

Importantly, the proposed patch-based processing makes our model input size-agnostic, and also introduces a light-weight generative diffusion modeling capability since the architecture can be based on a simpler backbone network for restoration at lower patch resolutions.
This way we extend the practicality of state-of-the-art diffusion model architectures with large computational resource demands in terms of the number of parameters and memory requirements during training and inference.
Our novel patch-based processing technique currently enables restoration of images on a single GPU with as little as 12GB memory.
Our approach also eliminates the restriction for the diffusion model backbone to have a fully-convolutional structure to be able to perform arbitrary sized image processing, and therefore our model can benefit from widely used resolution-specific attention mechanisms~\cite{vaswani2017attention,wang2018non}.

Our empirical analyses are mainly grounded on default architectural choices and minimal parameter settings used in seminal diffusion modeling works~\cite{Ho:2020,song2021ddim}.
By incorporating novel methods that improve diffusion models in terms of better sample quality~\cite{choi2022perception} or faster sampling mechanisms~\cite{Kingma:2021}, we argue that quantitative results can further be improved on particular weather restoration problems.

\subsection{Limitations}

The main limitation of our approach is its comparably longer inference duration, with respect to the end-to-end image restoration networks which only require a single forward pass for processing.
To illustrate an empirical example, our WeatherDiff$_{64}$ model requires 20.52 seconds (wall-clock time) to restore an image of size $640\times432$ with $S=10$ on a single NVIDIA A40 GPU, whereas TransWeather requires 0.88 seconds.
Such timing specifications of our method also directly rely on the choice of algorithm hyper-parameters (e.g., a lower value of $r$ slightly increases image quality but also the inference times), and implementation efficiency.

Another natural limitation of our model is its inherently limited capacity to only generalize to the restoration tasks observed at training time.
While we effortlessly enable multi-weather restoration by using image pairs from multiple corruptions at training time, this still does not qualify our generative model to conditionally generalize to unseen corruptions (e.g., poor lighting conditions).
Nevertheless, this natural limitation is also present in all recent studies that aim to tackle the problem of multi-weather image restoration~\cite{Li:2020CVPR,chen2022learning,Valanarasu:2022CVPR}.

\section*{Acknowledgments}

This work has been supported by the ``University SAL Labs" initiative of Silicon Austria Labs (SAL).

\bibliographystyle{IEEEtran}

\begin{IEEEbiography}[{\includegraphics[width=1in,height=1.25in,clip,keepaspectratio]{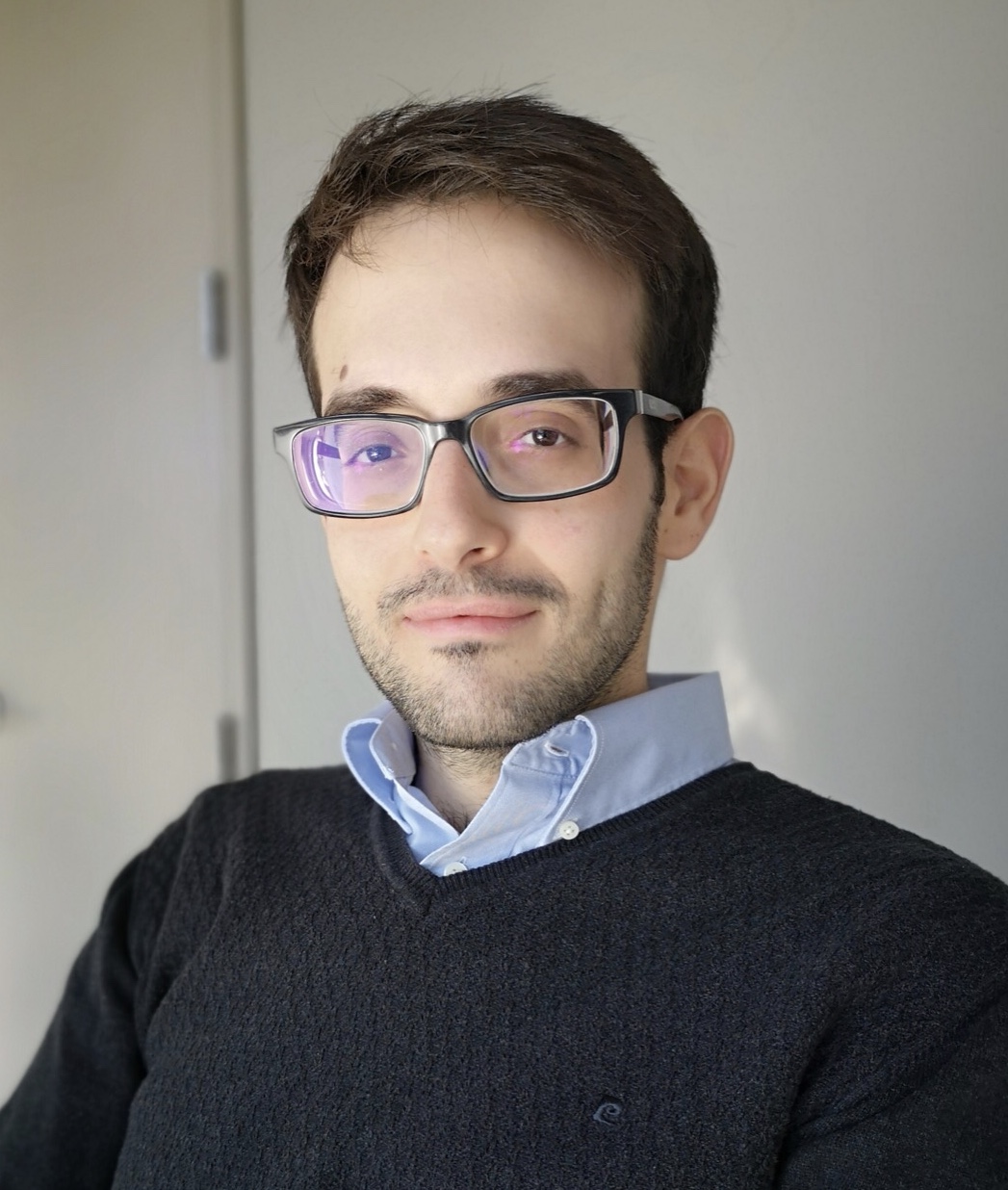}}]{Ozan \"{O}zdenizci} is a postdoctoral researcher at the Institute of Theoretical Computer Science, Graz University of Technology, Graz, Austria. He is also jointly affiliated with the TU Graz - SAL Dependable Embedded Systems Lab at Silicon Austria Labs.
He received his PhD in electrical engineering at Northeastern University, Boston, MA, USA, in 2020.
He received his BSc and MSc degrees from Sabanc{\i} University, Istanbul, Turkey, in 2014 and 2016 respectively. 
He previously held research stays at the Max Planck Institute for Intelligent Systems, T\"{u}bingen, Germany and Mitsubishi Electric Research Laboratories, Cambridge, MA, USA. 
His research interests are primarily in the domain of robust and reliable machine learning, and statistical signal processing with biomedical applications.
\end{IEEEbiography}

\begin{IEEEbiography}[{\includegraphics[width=1in,height=1.25in,clip,keepaspectratio]{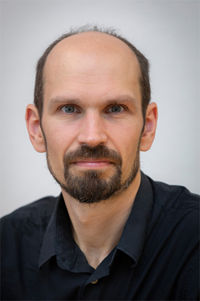}}]{Robert Legenstein} is currently a professor at the Department of Computer Science, TU Graz and head of the Institute of Theoretical Computer Science.
He received his PhD in computer science from Graz University of Technology, Graz, Austria, in 2002. In 2010 he got his Habilitation (venia docendi) for neuroinformatics. 
Robert Legenstein serves as action editor for Transactions on Machine Learning Research and has served as associate editor of IEEE Transactions on Neural Networks and Learning Systems. Robert Legenstein is a board member of the Austrian Society for Artificial Intelligence.
His primary research interests are learning in models for biological networks of neurons and neuromorphic hardware, brain-inspired machine learning, and deep learning.  
\end{IEEEbiography}

\end{document}